\RequirePackage{fix-cm}
\documentclass[12pt]{article}
\PassOptionsToPackage{table,dvipsnames}{xcolor}
\usepackage[left=2.5cm,
right=2.5cm,
top=2.3cm,
bottom=2.3cm,
headheight=20pt,
headsep=10pt,
footskip=25pt,
letterpaper]{geometry}
\usepackage[utf8]{inputenc}
\usepackage[T1]{fontenc}
\usepackage[english]{babel}
\usepackage{amsmath,amsfonts,amssymb,amsthm,thmtools}
\usepackage{graphicx}
\usepackage{hyperref}
\usepackage{fancyhdr}
\usepackage[normalem]{ulem}
\usepackage{graphicx}
\usepackage{stfloats}
\usepackage{wrapfig}
\usepackage{epstopdf}
\usepackage{cleveref}
\usepackage{subfloat}
\usepackage{subcaption}
\usepackage{xspace}
\usepackage{enumitem}
\usepackage{listings}
\usepackage{titlesec}
\usepackage{etoolbox}
\usepackage{setspace}
\usepackage{changepage}
\usepackage{etoolbox}
\usepackage{multirow}
\usepackage{booktabs}
\usepackage{tabularx}
\usepackage{wrapfig}
\usepackage{svg}
\usepackage[percent]{overpic}
\usepackage[round]{natbib}
\usepackage[colorinlistoftodos, shadow,color=blue!30!white
]{todonotes}
\usepackage{xpatch}
\usepackage{siunitx}

\fancypagestyle{first}{\fancyfoot[R]{\small\thepage}}

\setlist[itemize]{leftmargin=1em,itemsep=0ex,topsep=0ex}
\titlespacing*{\paragraph}{0pt}{0ex plus .1ex}{1ex}
\titlespacing*{\section}{0ex}{2.3ex plus .3ex minus .0ex}{.6ex plus .3ex minus .2ex}
\titlespacing*{\subsection}{0ex}{1.5ex plus .3ex minus .5ex}{.4ex plus .2ex minus .1ex}
\titlespacing*{\subsubsection}{0ex}{1.2ex plus .3ex minus .3ex}{.3ex plus .2ex minus .2ex}

\xapptocmd\normalsize{%
\abovedisplayskip=.8em plus .2em minus .2em
\belowdisplayskip=.6em plus .1em minus .1em
\abovedisplayshortskip=.8em plus .2em minus .2em
\belowdisplayshortskip=.6em plus .1em minus .1em
}{}{}

\setcitestyle{numbers}
\renewcommand{\cite}[1]{\citep{#1}}

\definecolor{mydarkblue}{rgb}{0.0,0.15,0.7}
\hypersetup{%
colorlinks=true,
linkcolor=mydarkblue,
citecolor=mydarkblue,
filecolor=mydarkblue,
urlcolor=mydarkblue}

\makeatletter
  \renewcommand{\maketitle}{%
    \begingroup
      {\centering\LARGE\@title\par}%
      \vskip 1em
      \centering
      \begin{tabular}[t]{@{}c@{}}\strut\@author\strut\end{tabular}%
      \vskip 0.3in minus 0.1in
    \endgroup
  }
\makeatother

\newif\ificlrstyle

\usepackage{fix-cm}
\usepackage[T1]{fontenc}
\usepackage[utf8]{inputenc}
\usepackage{amsmath,amssymb}
\usepackage{graphicx}
\usepackage{float}
\usepackage{wrapfig}
\usepackage[table,dvipsnames]{xcolor}
\usepackage{array,tabularx,booktabs,makecell,multirow}
\usepackage{microtype}
\usepackage{url}
\usepackage{hyperref}
\usepackage[ruled,linesnumbered]{algorithm2e}

\definecolor{TableHeader}{HTML}{DAD8E2}
\definecolor{TableStripe}{HTML}{F2F2F2}
\definecolor{TableOurs}{HTML}{ECECF7}
\definecolor{ChangeUp}{HTML}{B95843}
\definecolor{ChangeDown}{HTML}{248892}
\definecolor{RuleColor}{HTML}{353535}
\definecolor{LinkColor}{HTML}{244E74}
\hypersetup{colorlinks=true,citecolor=LinkColor,linkcolor=LinkColor,urlcolor=LinkColor}
\newcommand{\method}{\textsc{StateComp}}
\newcommand{\gain}[1]{\,{\scriptsize\textcolor{ChangeUp}{(#1)}}}
\newcommand{\best}[1]{\textbf{#1}}
\newcommand{\second}[1]{\underline{#1}}
\newcommand{\na}{\textnormal{n/a}}
\newcommand{\tablefont}{\fontsize{8.5}{10.2}\selectfont}
\newcommand{\tabsetup}{%
  \centering\tablefont
  \setlength{\tabcolsep}{4.5pt}%
  \renewcommand{\arraystretch}{1.14}%
  \arrayrulecolor{RuleColor}}

\newcolumntype{Y}{>{\centering\arraybackslash}X}
\definecolor{TableHeader}{RGB}{226,224,234}
\definecolor{TableStripe}{RGB}{246,246,246}
\definecolor{TableOurs}{RGB}{237,236,247}

\definecolor{UpColor}{RGB}{215,92,72}
\definecolor{DownColor}{RGB}{28,145,158}

\newcommand{\up}[1]{%
  \hspace{1.2pt}{\scriptsize\textcolor{UpColor}{$\uparrow$#1}}%
}
\newcommand{\dn}[1]{%
  \hspace{1.2pt}{\scriptsize\textcolor{DownColor}{$\downarrow$#1}}%
}
\newcommand{\RouterThreshold}{0.60}
\newcommand{\SpanThreshold}{3}
\newcommand{\SourceThreshold}{1{,}000}
\newcommand{\StateBudget}{5{,}120}

\newcommand{\paperfigure}[2][0.96\linewidth]{%
  \IfFileExists{#2}{\includegraphics[width=#1]{#2}}{%
    \fbox{\parbox[c][30mm][c]{0.90\linewidth}{\centering
      Figure file required: \texttt{\detokenize{#2}}}}}}
\SetAlgoNoLine
\DontPrintSemicolon
\SetNlSty{textnormal}{}{:\ }
\SetKwInput{KwRequire}{Input}
\SetKwInput{KwEnsure}{Output}
\SetKw{Return}{return}
\SetAlCapNameFnt{\bfseries}
\SetAlgoCaptionSeparator{\ }
\title{StateComp: Learning When to Compress History in Long Horizon Agents}
\author{%
  \mbox{Mingxuan Wang\textsuperscript{1}}\quad \mbox{Hongyue Chen\textsuperscript{1}}\quad \mbox{Yinglong Guo\textsuperscript{1}}\quad \mbox{Fei Luo\textsuperscript{1}}\quad \mbox{Chao Ning\textsuperscript{1}}\\[2pt]
  \mbox{Bo Wang\textsuperscript{1}}\quad \mbox{Guorun Yao\textsuperscript{1}}\quad \mbox{Yanbiao Ma\textsuperscript{2,*}}\quad \mbox{Jungong Han\textsuperscript{3,*}}\\[4pt]
  \textsuperscript{1}TierFlow Team\\
  \textsuperscript{2}Gaoling School of Artificial Intelligence, Renmin University of China\\
  \textsuperscript{3}Tsinghua University\\[3pt]
  \textsuperscript{*}Corresponding authors.\quad \href{mailto:ybma1998@ruc.edu.cn}{\texttt{ybma1998@ruc.edu.cn}}%
}
\hypersetup{pdfauthor={Mingxuan Wang, Hongyue Chen, Yinglong Guo, Fei Luo, Chao Ning, Bo Wang, Guorun Yao, Yanbiao Ma, Jungong Han}}

\date{}
\hypersetup{colorlinks=true,linkcolor=mydarkblue,citecolor=mydarkblue,urlcolor=mydarkblue}
\usepackage{team-template}
\renewcommand{\TeamPaperID}{STATECOMP / LONG-HORIZON AGENTS}
\renewcommand{\TeamShortTitle}{StateComp}
\begin{document}
\pagestyle{fancy}
\maketitle
\thispagestyle{first}

\begin{abstract}
Long-horizon agents continuously accumulate interaction history during task execution, yet the importance of past interactions changes as the agent state evolves. Existing context management methods largely compress history based on fixed windows, periodic schedules, or current relevance, overlooking a more fundamental question: when has a past interaction become safe to replace? Premature compression may remove information still needed for future actions, while overly conservative retention leads to substantial context overhead. To address this, we propose \textbf{State Conditioned Compression (\method{})}, a framework that determines when historical interactions can be safely compressed according to the current agent state. \method{} constructs KEEP and READY supervision through a two-stage annotation procedure and trains an imbalance-aware router on hidden representations from a frozen language model. A bounded state representation further reduces the cost of evaluating long histories, while adjacent READY interactions are grouped into continuous spans and replaced with compact summaries during execution. Experiments on WorkBuddyBench show that \method{} reduces total agent and summarization tokens by $52.27\%$ while maintaining task performance, and achieves a $12.67\times$ speedup in representation extraction.

\end{abstract}

\section{Introduction}
\label{sec:introduction}

Long horizon agents continuously accumulate task instructions, intermediate
reasoning, tool calls, environment observations, and unsuccessful attempts
throughout execution \citep{yao2022react,shinn2023reflexion,wang2023voyager,liu2024agentbench}. As the trajectory grows, each new step may require the
model to process an increasingly long interaction history \citep{liu2024lost,bai2024longbench,bai2025longbench}, resulting in
substantial token and inference costs \citep{jiang2023llmlingua,jiang2024longllmlingua,li2023compressing}. However, not every past interaction
needs to remain in its original form throughout the entire task. Some
interactions continue to determine future actions \citep{wu2024longmemeval,maharana2024evaluating}, while others have already
served their purpose and their essential information has been absorbed by
later observations or conclusions. The central challenge is therefore not
simply how to compress a long prompt, but rather which past interactions
have become obsolete at the current agent state and can now be safely
compressed during execution.

This decision is inherently state dependent. The same historical interaction
may have very different retention requirements at different stages of
execution. For example, an error trace can be indispensable while the agent is
diagnosing a failure, but its complete details may no longer be necessary once
the cause has been identified and the relevant conclusion has been recorded.
In contrast, an early task constraint may remain important until the end of
the trajectory. Compression opportunities are therefore distributed
nonuniformly throughout agent history and cannot be determined reliably from
age, position, or a fixed context window alone. At each current state, the
agent must reconsider which earlier interactions still require their complete
details and which have already become replaceable.

This problem is also strongly asymmetric. In our supervision data, most
historical interaction and state pairs remain KEEP, while only a small
fraction are labeled READY. Keeping an interaction that could already be
compressed mainly incurs additional context cost. Compressing an interaction
that is still needed, however, may permanently remove information required by
subsequent actions. This distinction is particularly important in our setting
because removed raw interactions are not recovered through external
retrieval. We therefore construct compression supervision with a two stage
annotation procedure and train the router with an imbalance aware objective,
favoring precise identification of truly compressible history over aggressive
compression coverage.

Motivated by these observations, we introduce \textbf{State Conditioned Compression (\method{})}, a framework that determines when historical interactions can be
safely compressed according to the current agent state during long-horizon execution. The key idea is to use
the current agent state to repeatedly reassess previous interactions and to
separate the prediction that some history has become compressible from the
decision to execute an actual compression operation. The main contributions
of this work are as follows:

{
\setlength{\leftmargini}{1.2em}
\setlength{\itemsep}{3pt}

\begin{itemize}
    \item \textbf{State dependent compression supervision.}
    We formulate history compression as a joint decision over a past
    interaction and the current agent state, and construct KEEP and READY
    supervision through a two stage annotation procedure that identifies
    supported compression boundaries.

    \item \textbf{Dynamic compression routing from hidden states.}
    We train an imbalance aware router on hidden representations from a
    frozen language model to predict whether each historical interaction
    can be safely compressed during online inference. A bounded state representation further reduces
    representation extraction cost.

    \item \textbf{Stateful online compression execution.}
    We group adjacent READY interactions into continuous spans and summarize
    only sufficiently large candidates. Committed summaries directly replace
    the corresponding raw history and become part of the effective context
    for subsequent agent actions and routing decisions throughout later execution.
\end{itemize}
}

We evaluate \method{} on WorkBuddyBench \citep{team2026tencent}. Across the full set of 260 tasks,
total agent and summarization tokens decrease from 698.17M to 333.24M, a
reduction of $52.27\%$, while mean reward changes from $0.6987$ to $0.7026$.
Additional experiments on Eval40, representation efficiency, and multiple
acting models further examine the quality and context cost tradeoff of
\method{} across different agent settings.

\section{Related Work}
\label{sec:related_work}

\subsection{Prompt Compression}
LLMLingua-2 formulates extractive prompt compression as token classification
\citep{pan2024llmlingua}. This provides a way to shorten an input while
retaining selected information \citep{jiang2023llmlingua,jiang2024longllmlingua,li2023compressing,chevalier2023adapting,mu2023learning}. In an agent setting, prompt compression and
compression timing address different parts of history management \citep{ge2023context,xu2024recomp,cheng2024xrag,zhang2025long}. A
representation of what should remain in a shortened input does not, by
itself, specify the point in task execution at which the original details
can be replaced. \method{} focuses on this temporal decision at the level
of complete interactions. Its router selects the history that is ready for
compression, while a separate summarizer constructs the content that remains.
The selection target is defined jointly by a historical interaction and the
current state rather than by the interaction alone.

\subsection{Context Management for Long Horizon Agents}
SelfCompact gives agents a compaction tool and guidance on when to use it,
while ACON studies context compression for long horizon execution
\citep{li2026self,kang2025acon}. Self-GC, LRE, CoMem, and SAM explore
complementary approaches to context management and memory
\citep{hao2026self,jahan2026learning,zhang2026comem,hu2026sam,packer2023memgpt,park2023generative,zhong2024memorybank,chhikara2025mem0,kang2025memory,xu2026mem,gutierrez2024hipporag}.
These works place context management within the continuing execution of an
agent rather than treating history only as a fixed input. \method{} uses a
separate router trained on supported compression boundaries. Its prediction
and execution decisions are distinct: a positive router prediction can
contribute to a candidate span without immediately triggering summarization.
Once a summary is committed, later actions and routing use the updated
effective history. This makes the timing of replacement, the size of the
selected region, and the information retained by the summary separate parts
of the compression procedure.

\subsection{Adaptive Selection of Historical Information}
SWE-Pruner selects useful context for coding agents, and PACE adapts
historical context using next step relevance
\citep{wang2026swe,wei2026pace}. Sculptor equips agents with active context
management tools, while ACM also studies context management over long
horizon tasks \citep{li2026sculptor,li2026acm}. These approaches are related
to selecting information for ongoing decisions. The target in \method{} is
specifically whether an interaction still needs to remain in its original
form at the current checkpoint. The annotation checks evidence already
available in the prefix and uses future dependencies only to reject unsafe
candidates. Selected raw spans are replaced by summaries that remain in the
effective history. The online controller does not subsequently retrieve the
removed originals, so both the decision to compress and the content of each
replacement matter to later execution. This distinction separates replaceability from conventional relevance
selection: an interaction may remain related to the current task while no
longer requiring its original high-detail representation during subsequent execution.

\section{Supervision for State Dependent Compression}
\label{sec:safe_compression}

A past interaction does not become unnecessary simply because it is old.
For example, an error trace may determine the next debugging action, but its
full details may become replaceable once the failure has been diagnosed and
the relevant conclusion has been recorded. We define compression with respect
to the information available at the current agent state.

\subsection{Compression Depends on the Current State}
\label{sec:state_dependent_compression}

We use a complete interaction as the basic unit,
$S_i=(R_i,A_i,O_i)$, containing reasoning, the assistant action, and its
observation. Compression respects complete interaction boundaries so that tool
calls and their responses remain paired. Before interaction $k$, the completed
history is $P_k=(S_1,\ldots,S_{k-1})$. For $i<k$, the label $y_{i,k}=0$
means that $S_i$ should remain in full, whereas $y_{i,k}=1$ means that its
original details can be replaced by an accurate summary at that state. Importantly, this label is not an intrinsic property of $S_i$ alone: the same
interaction may transition from KEEP to READY as later observations resolve
uncertainty, establish conclusions, or make its original details redundant.

The label is a property of the interaction and the current state together.
Two interactions at similar positions can have different retention needs.
An early task constraint may remain necessary throughout execution, while an
intermediate search result may be superseded by a confirmed finding. The
annotation therefore checks whether necessary information would remain
available after removing the original details, rather than using age as a
proxy for relevance.

\subsection{The Earliest Safe Compression Point}
\label{sec:safe_boundary}

For each $S_i$, we identify the earliest checkpoint $t_i^*$ from which its
full details are judged unnecessary along the annotated continuation. The
resulting labels are
\begin{equation}
 y_{i,k}=\begin{cases}
  0, & k<t_i^*,\\
  1, & k\geq t_i^*.
 \end{cases}
 \label{eq:safe_boundary}
\end{equation}
If no supported point is found, we set $t_i^*=\infty$ and retain the
interaction throughout the trajectory. This boundary summarizes the
annotation for an observed trajectory; it is not a guarantee for every
possible future continuation. Using the earliest supported boundary also avoids treating eventual
compressibility as evidence that an interaction could have been removed
earlier in the trajectory.

This formulation treats compression as a state-dependent transition rather
than a static importance judgment. An interaction becomes compressible only
when the current prefix provides sufficient evidence to replace its raw form.

A positive decision requires that information still needed later is already
available elsewhere in the retained history, or that the original detail is no
longer needed. The final dataset contains 244,526 interaction and state pairs,
of which 15,633 are positive. In total, 1,832 historical interactions receive a
supported compression point. Complete statistics appear in
Appendix~\ref{app:experimental_details}.

\subsection{Two Stage Annotation with Supporting Evidence}
\label{sec:boundary_annotation}

Direct annotation from a complete trajectory can confuse eventual resolution
with information already available at an earlier checkpoint. We therefore
require positive evidence to come from $P_k$. Future interactions may reveal
that deleting $S_i$ would remove a detail needed later, but future success or
a later summary cannot establish that an earlier deletion was safe.

Stage 1 independently annotates each checkpoint. It records task progress,
observation complexity, expected next subtask requirements, the remaining
utility of the target interaction, and any supporting evidence already present
in the prefix. Each record includes a reason, evidence locations, an initial
KEEP or READY judgment, and any future dependency that vetoes a positive
judgment. Decisions at one checkpoint are not inherited by another.

Stage 2 uses these Stage 1 records as its input. For a fixed historical
interaction, it reviews the initial decisions across checkpoints together with
the original evidence, checks proposed compression points again, and finds
the earliest supported boundary. This separates local evidence collection
from final boundary determination. It also exposes inconsistent initial
judgments instead of propagating the first READY label through all later
states. Unresolved cases remain KEEP. This decomposition is intended to make positive labels depend on explicit
state-local evidence rather than on a single holistic judgment over the full
trajectory.

The procedure treats the two errors differently. Retaining a replaceable
interaction mainly increases context cost. Removing information that remains
necessary can harm subsequent actions, especially when raw history cannot be
retrieved later. We therefore require stronger evidence for a positive label
and favor precision when selecting a router operating point.

\section{StateComp Methodology}
\label{sec:method}

\begin{figure}[t]
 \centering
 \paperfigure[0.98\linewidth]{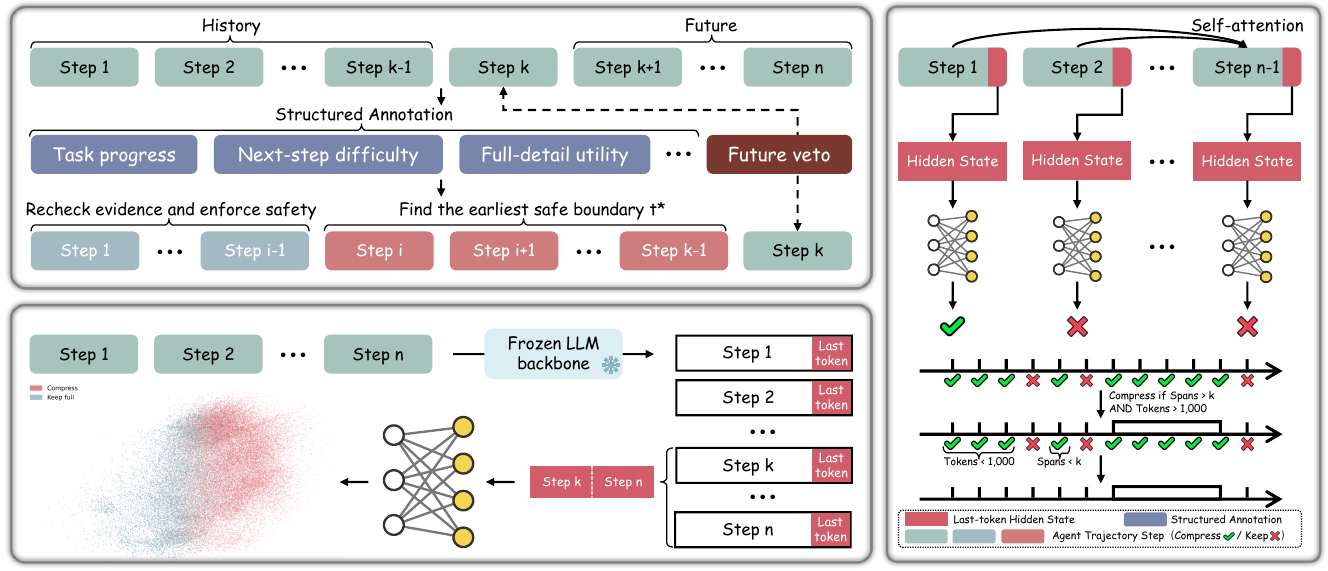}
 \caption{\textbf{Overview of \method{}.}
 Structured annotations and a second boundary review produce supervision.
 A router learns from hidden states of a frozen representation model.
 Online predictions are combined with span length and token gates before
 summaries replace raw interactions. Historical and current state features
 are used jointly, as described in Section~\ref{sec:hidden_state_question}.}
 \label{fig:method_overview}
 \vspace{-10pt}
\end{figure}

\method{} uses the current state to reconsider which historical interactions
must remain in full. It makes predictions at interaction granularity,
summarizes eligible continuous spans, and updates the history used by later actions. A central design choice is to separate readiness prediction from
compression execution: identifying a replaceable interaction does not
by itself imply that an immediate rewrite is worthwhile. This separation allows the controller to accumulate local compression
opportunities before committing a larger rewrite, reducing unnecessary
summary calls and avoiding fragmented updates to the active history.
Figure~\ref{fig:method_overview} connects labeling, router training,
and online compression throughout the subsequent task execution.

\subsection{Hidden-State Representations and Router Training}
\label{sec:hidden_state_question}
\label{sec:bounded_representation}

The online agent cannot access future interactions or the annotated boundary.
We instead read representations from a frozen language model. In the full
context reference, one forward pass over $P_k$ provides a hidden vector $h_i$
at the final token of each historical interaction and a current state vector
$q_k$ at the final prompt position. The router uses the historical vector
together with the current state to predict $y_{i,k}$ at each checkpoint.

This formulation does not require an earlier token to attend to later tokens.
With a causal attention mask, appending new interactions does not by itself
change an earlier hidden vector. State dependence is supplied by $q_k$ and
its combination with the target interaction, or by a separately constructed
input that contains both the target and current context. Thus, the same historical interaction can receive different
router scores as the current state changes. Section~\ref{sec:method} describes
the bounded implementation used to reduce representation extraction cost.

Full context extraction is a useful reference but becomes expensive as history
grows \citep{kwon2023efficient,xiao2024efficient,zhang2023h2o,jiang2024minference,xiao2024infllm}. The recorded trajectories include inputs of approximately 128K tokens
and a largest input of 203,675 tokens. Repeatedly processing these histories
can make the representation model itself a substantial source of overhead \citep{liu2023scissorhands,li2024snapkv,cai2024pyramidkv,tang2024quest}.

The practical construction uses a separate bounded view for each historical
interaction under evaluation. It combines the target interaction with the
current state and a limited amount of effective history. The frozen model
reads this view, and the final prompt position supplies a compact readout for
the decision. Every still present raw interaction remains eligible for
assessment. Bounded input construction changes the context used to compute
the features, not the range of historical interactions considered by the router. Thus, bounding the representation cost does not impose a recency-based
restriction on which surviving historical interactions may be reconsidered.

The representation budget is at most $\StateBudget$ tokens per view.
Recent4 is the reference recent context construction in the reported bounded
ablation. We distinguish its context window $K$ from the compression span
threshold $\kappa$ below. With $N$ historical targets, candidate specific
construction involves $N$ short input sequences, which may be batched. The
per-view token bound does not make the complete checkpoint cost independent
of $N$.

For each remaining historical interaction, the router combines the target
representation, current state, and available historical features to produce a
score $p_{i,k}$. Features come only from the effective prefix. Aggregating
historical vectors within that prefix is a prediction operation and does not
restore previously removed raw content. The router is trained on the labels
from Section~\ref{sec:safe_compression} with the representation model frozen
throughout all training.
Training variants and the grouped selection protocol are documented in
Appendix~\ref{app:router_protocol}.

\subsection{Interaction Routing and Span Execution Gates}
\label{sec:compression_spans}

A score at or above $\tau$ selects an interaction as a compression candidate.
The controller then merges adjacent selected interactions into continuous
spans. A KEEP interaction breaks a span; it is not included merely to connect
two selected regions. A positive router decision does not immediately trigger
a summary call. For a candidate span $B=[s,e]$, the execution rule is
\begin{equation}
 \min_{i\in B}p_{i,k}\geq\tau,\qquad
 |B|>\kappa,\qquad
 \operatorname{Tokens}(B)\geq B_{\min},
 \label{eq:execution_gates}
\end{equation}
with $\tau=\RouterThreshold$, $\kappa=\SpanThreshold$, and
$B_{\min}=\SourceThreshold$. Thus, an eligible span contains at least four
complete interactions and at least 1,000 source tokens. The token condition
refers to the source length, not a guaranteed number of saved tokens.

These gates address the cost of executing compression. The score threshold controls prediction confidence, while the span-length and
source-token conditions control whether a predicted opportunity is large
enough to justify an actual summarization operation. Summarizing tiny
fragments can spend more tokens on the auxiliary call than it saves in the
active history. Combining adjacent interactions also gives the summarizer a
more coherent source and avoids a succession of fragmented replacements. The execution gates therefore convert local readiness predictions into
larger operational decisions, separating semantic replaceability from the
economic cost of performing a rewrite. Importantly, rewriting earlier history can invalidate KV reuse from the first
changed position onward. Prefix caching depends on an unchanged prefix. Deferring small rewrites is intended to reduce repeated
cache invalidation and reconstruction. It does not preserve the KV states of
the rewritten suffix or guarantee a higher cache hit rate without measurement.

Each eligible span produces one summary containing facts, constraints,
locations, confirmed conclusions, failure causes, and unresolved items that
remain useful. The router determines which original details may leave the
active context; the summarizer determines what the replacement retains.
The controller commits a replacement only when it preserves interaction
structure and actually shortens the effective history. Otherwise, it keeps the
original span. Appendix~\ref{app:online_algorithm} specifies this procedure.

\subsection{Updating the Effective Agent History}
\label{sec:stateful_update}

After a summary replaces a raw span, the resulting effective history becomes
the context used for the next agent action. Later state features and router
predictions are computed from this updated history rather than from an
untouched copy of the original trajectory. Remaining interactions are therefore
reconsidered at every checkpoint under the newly formed agent state. Compression is therefore a closed-loop process: each committed replacement
changes the context on which subsequent routing decisions are made throughout later execution.

\begin{wrapfigure}{r}{0.50\linewidth}
    \vspace{-0.8em}
    \centering
    \includegraphics[width=\linewidth]{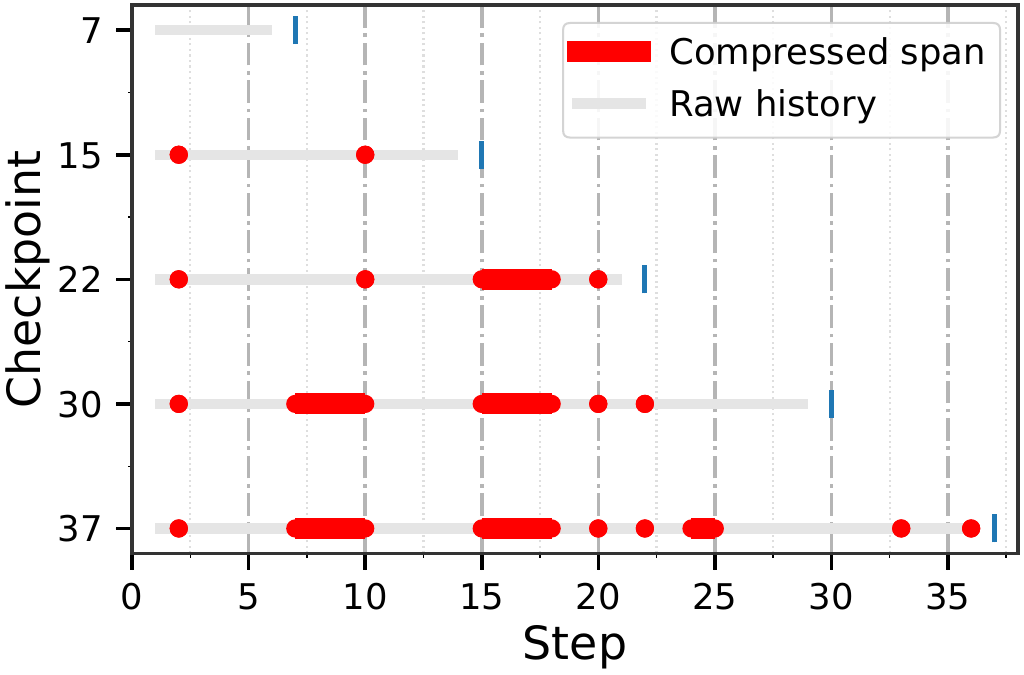}
    \vspace{-18pt}
    \caption{Compression dynamics across checkpoints in an example trajectory.}
    \label{fig:compression_example}
    \vspace{-0.7em}
\end{wrapfigure}

Figure~\ref{fig:compression_example} illustrates this process on one
trajectory. Early checkpoints retain most of the original history. As
execution proceeds, additional continuous regions become eligible for
compression and are replaced, while interactions that still require their
original details remain in the active history.

This stateful update is important because a compression action changes the
context available to all subsequent decisions. A previously selected
interaction is therefore not treated as a permanent authorization for future
compression. Instead, the router scores the remaining history again from the
current effective state.

The system retains the original trajectory separately for auditing, but this
canonical record is not used as an external retrieval store by the online
controller. The effective history contains only retained raw interactions and
committed summaries. If routing or final validation fails, the controller
returns the previous valid effective history. A failed summarization
proposal leaves its source span unchanged.

\section{Experiments}
\label{sec:experiments}

We evaluate task reward and model token consumption on complete agent tasks.
The full benchmark measures the overall effect, the fixed Eval40 comparison
examines alternative context strategies, and cross-model runs characterize
how the quality and token tradeoff changes with the acting model.

\subsection{Experimental Setup}
\label{sec:system_setup}

We first examine how labeled compression opportunities evolve, then evaluate
prediction on held out trajectories. The strict router study uses frozen
Qwen2.5-7B representations and five grouped folds. Each rotation has 200
training, 50 validation, and 50 test trajectories. All pairs from one
trajectory stay in the same split. Configuration and threshold selection use
training and validation data rather than test labels in every fold.

WorkBuddyBench contains 260 tasks: 80 Code, 50 Office, 60 Security, and 70 Web
\citep{team2026tencent}.
The full comparison uses DeepSeek-V4-Flash \citep{xu2026deepseek} on the same
260 tasks. Eval40 contains ten fixed tasks from each domain. All method
and cross-model comparisons on Eval40 use this same 40-task set. We report mean reward
multiplied by 100 in tables, with unscaled reward in the text. Total tokens
include recorded agent and summary input and output tokens. Cached input and
reasoning output are subsets of those totals and are not added again. Local
representation cost is reported separately in
Table~\ref{tab:bounded_representation}.

The Full260 and Eval40 evaluations use the compression rule in
Section~\ref{sec:compression_spans}, with $\tau=\RouterThreshold$,
$\kappa=\SpanThreshold$, and $B_{\min}=\SourceThreshold$ tokens.
The additional published baselines use the same Eval40 token accounting.
Operating settings and the fixed Eval40 task identifiers are provided in
Appendix~\ref{app:parameter_studies}.

\begin{table*}[t]
\centering
\caption{
\textbf{Efficiency and quality of bounded state representations.}
Max. tokens reports the largest recorded input; runtime and memory are measured on the same seven paired states.
}
\label{tab:bounded_representation}

\small
\setlength{\tabcolsep}{6pt}
\renewcommand{\arraystretch}{1.12}

\begin{tabular}{lccccc}
\toprule

\textbf{Method}
&
\textbf{Max. tokens}
&
\textbf{Wall clock (s)}
&
\textbf{Peak GPU (GiB)}
&
\textbf{Workspace (GiB)}
&
\textbf{Test F1}
\\

\midrule

Full history
&
203,675
&
76.40
&
29.426
&
11.867
&
0.7000
\\

\rowcolor{TableOurs}
\textbf{Bounded}
&
\textbf{5,120}\,\gain{-97.49\%}
&
\textbf{6.03}\,\gain{12.67$\times$}
&
\textbf{18.620}\,\gain{-36.72\%}
&
\textbf{1.062}\,\gain{-91.05\%}
&
\textbf{0.7818}\,\gain{+0.08}
\\

\bottomrule
\end{tabular}
\vspace{-15.65pt}
\end{table*}
\subsection{Hidden-State Prediction and Representation Efficiency}
\label{sec:hidden_state_analysis}
\label{sec:compression_dynamics}
\label{sec:hidden_state_structure}

\begin{wrapfigure}{r}{0.50\linewidth}
    \vspace{-0.6em}
    \centering
    \includegraphics[width=\linewidth]{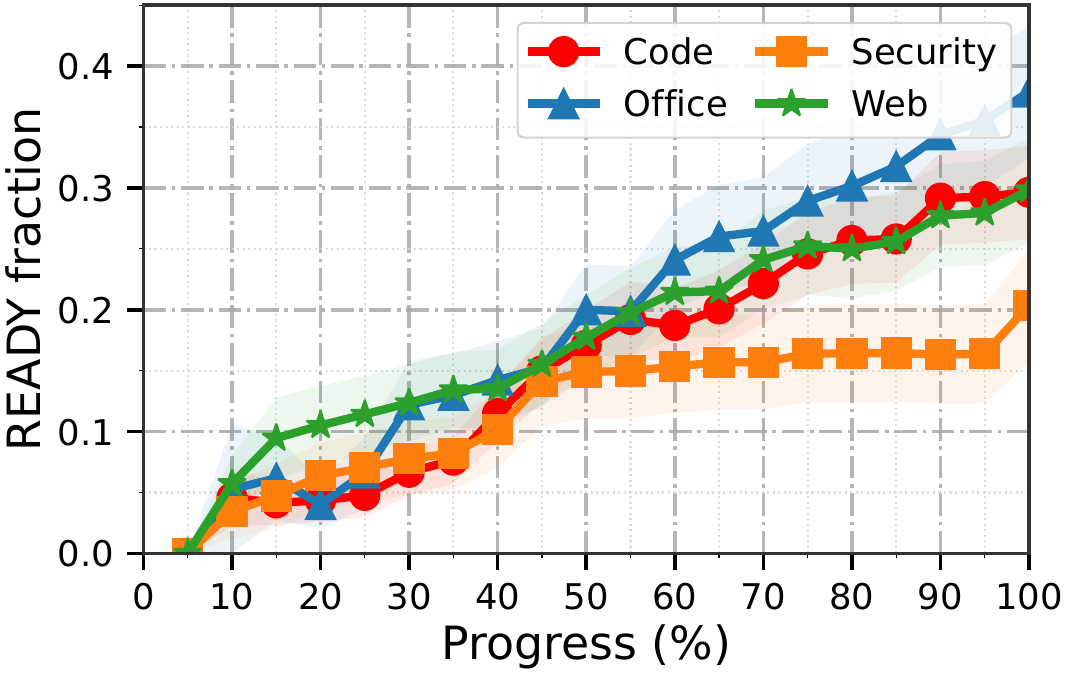}
    \vspace{-20pt}
    \caption{
    Compression opportunities over different trajectory stages.
    }
    \label{fig:readiness_progress}
    \vspace{-0.7em}
\end{wrapfigure}

Figure~\ref{fig:readiness_progress} shows that compression opportunities
emerge progressively during task execution. Across all four domains, only a
small fraction of past interactions are labeled READY near the beginning of a
trajectory, while the fraction generally increases as execution proceeds.
This pattern indicates that historical interactions do not have a fixed
retention requirement. Instead, their compression status changes as the agent
accumulates new evidence and completes intermediate subtasks.

The trend motivates state conditioned selection rather than compression based
only on age or a fixed context window. At the same time, the four domains
follow different trajectories, suggesting that task progress alone is not a
sufficient decision rule in practice. This variation further supports conditioning on the observed agent state
rather than applying a globally shared compression schedule across tasks. The router therefore conditions each compression
decision on the current agent state and the target historical interaction.
Additional analyses of compression delay, trajectory variation, and trajectory
length are reported in Appendix~\ref{app:compression_dynamics}.

We next test whether these state dependent compression decisions can be
predicted from hidden representations. Under grouped cross trajectory
evaluation, a representative configuration obtains AUROC $0.7096$ and
PR AUC $0.1736$. The result indicates that frozen hidden states contain
predictive information about whether the full details of a historical
interaction should still be retained. The signal is therefore informative but not sufficient for unrestricted
compression, motivating the conservative operating policy used in the
online controller.

% Added from the right panel of the existing fig04.pdf.
\Needspace{210pt}
\begin{wrapfigure}{r}{0.50\linewidth}
    \vspace{-10pt}
    \centering
    \includegraphics[width=\linewidth]{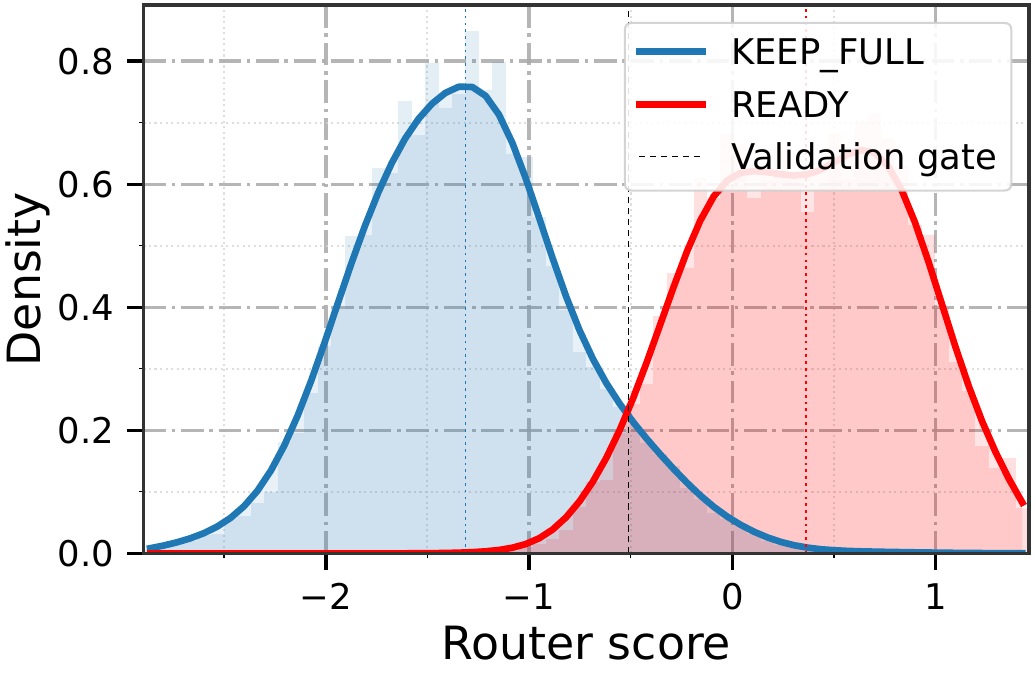}
    \vspace{-25pt}
    \caption{
Router score distributions.
}
    \label{fig:router_score_distribution}
    \vspace{-15pt}
\end{wrapfigure}
The prediction problem is intentionally asymmetric. Retaining an interaction
that could already be compressed mainly increases context cost, whereas
compressing an interaction too early can remove information needed by later
actions. We therefore use a precision oriented operating policy. A highly
selective configuration reaches $72.22\%$ external precision at $0.11\%$
recall. Full precision coverage curves, calibration results, and threshold
diagnostics are provided in Appendix~\ref{app:router_ablation}.

Figure~\ref{fig:router_score_distribution} shows the learned score distributions
for KEEP\_FULL and READY examples. This is a visualization of the trained
readout; the grouped evaluation above reports predictive performance.

Table~\ref{tab:bounded_representation} separates recorded context sizes from
paired runtime measurements. On seven states for which both implementations
complete, total extraction time falls from 76.40 to 6.03 seconds. Peak GPU
memory decreases by $36.72\%$ and forward workspace by $91.05\%$.
The largest full inputs do not complete in this microbenchmark and are not
used as timed paired samples. In the separate representation comparison, the
best reported bounded configuration has F1 $0.7818$, compared with $0.7000$
for the full context reference. Thus, the measured efficiency gain is not accompanied by lower prediction
quality in this comparison, although the bounded and full representations
should not be interpreted as information-equivalent. Detailed configurations and measurements are
in Appendix~\ref{app:representation_design}. Taken together, these results show that compression readiness is both
state varying and partially predictable from compact representations,
providing the basis for practical online routing.

\subsection{Task Performance and Context Cost}
\label{sec:system_results}
\label{sec:method_comparison}

Table~\ref{tab:domain_results} reports the complete comparison. Mean reward
changes from $0.6987$ to $0.7026$, while total agent and summary tokens fall
from 698.17M to 333.24M, a reduction of $52.27\%$. Average tokens per task
therefore fall from approximately 2.69M to 1.28M. The observed result is a
large reduction in transmitted and generated tokens with a small positive
change in average task reward. The main empirical benefit in this setting is therefore efficiency rather than a claim of
systematic reward improvement: the method removes substantial context while
keeping aggregate task quality close to the uncompressed execution.

Token reductions occur in all four domains, ranging from $38.89\%$ in Code
to $55.75\%$ in Office. Code, Office, and Web have
slightly higher reward, while Security changes from $0.4776$ to $0.4731$.
These results show consistent token reductions across domains, with small
domain-dependent changes in task reward. Detailed token components are provided in
Appendix~\ref{app:online_results}. The consistent token reductions across domains indicate that the observed
efficiency gain is not concentrated in a single task category in this evaluation.

\begin{table}[t]
\centering
\caption{
\textbf{Full WorkBuddyBench comparison on 260 tasks.}
Scores are mean reward multiplied by 100.
Tokens include the main agent and summary calls.
Colored changes show score differences relative to the baseline for all domains.
}
\vspace{+3pt}
\label{tab:domain_results}

\tabsetup
\setlength{\tabcolsep}{4.8pt}
\renewcommand{\arraystretch}{1.10}

\resizebox{\linewidth}{!}{%
\begin{tabular}{l|cccccc}

\Xhline{1.1pt}

\rowcolor{TableHeader}
\textbf{Method}
&
\textbf{Code}
&
\textbf{Office}
&
\textbf{Sec.}
&
\textbf{Web}
&
\textbf{Avg.}
&
\textbf{Tokens (M)}
\\

\Xhline{0.8pt}

Baseline
&
76.99
&
81.84
&
\best{47.76}
&
72.14
&
69.87
&
698.17
\\

\rowcolor{TableOurs}
\textbf{\method{}}
&
\best{77.12}\up{0.13}
&
\best{82.77}\up{0.93}
&
47.31\dn{0.45}
&
\best{73.14}\up{1.00}
&
\best{70.26}\up{0.39}
&
\best{333.24}
\\

\Xhline{1.1pt}

\end{tabular}}
\vspace{-10pt}
\end{table}

\begin{table*}[t]

\centering

\caption{
\textbf{Context management results on Eval40.}
Scores are reward multiplied by 100.
Small colored arrows show changes relative to the DeepSeek baseline.
Bold and underlined values indicate the best and second best results.
}
\vspace{+3pt}
\label{tab:method_comparison}

\renewcommand{\arraystretch}{1.08}
\setlength{\tabcolsep}{4.6pt}

\resizebox{\textwidth}{!}{%
\begin{tabular}{l|cccccc}

\Xhline{1.2pt}

\rowcolor{TableHeader}
\textbf{Method}
&
\textbf{Code}
&
\textbf{Office}
&
\textbf{Sec.}
&
\textbf{Web}
&
\textbf{Avg.}
&
\textbf{Tokens (M)}
\\

\Xhline{1.0pt}

Baseline
& 72.93
& 81.64
& 44.58
& 71.00
& 67.54
& 75.50
\\

\rowcolor{TableStripe}
Sliding Window ($K=5$)
& 66.81\dn{6.12}
& 67.43\dn{14.21}
& 32.32\dn{12.26}
& 59.00\dn{12.00}
& 56.39\dn{11.15}
& 118.08
\\

Sliding Window ($K=10$)
& 74.37\up{1.44}
& 73.05\dn{8.59}
& 40.07\dn{4.51}
& 61.00\dn{10.00}
& 62.12\dn{5.42}
& 79.39
\\

\rowcolor{TableStripe}
Sliding Window ($K=20$)
& \second{79.31}\up{6.38}
& \best{85.11}\up{3.47}
& 40.28\dn{4.30}
& 70.00\dn{1.00}
& 68.67\up{1.13}
& 80.82
\\

Periodic Summary ($n=3$)
& 65.44\dn{7.49}
& 67.66\dn{13.98}
& 26.32\dn{18.26}
& \best{82.66}\up{11.66}
& 60.52\dn{7.02}
& 88.57
\\

\rowcolor{TableStripe}
Periodic Summary ($n=5$)
& \best{83.60}\up{10.67}
& 75.74\dn{5.90}
& 40.03\dn{4.55}
& \second{81.32}\up{10.32}
& \second{70.17}\up{2.63}
& 94.85
\\

\hline

LLMLingua-2~\citep{pan2024llmlingua}
& 64.64\dn{8.29}
& 73.31\dn{8.33}
& 40.10\dn{4.48}
& 70.00\dn{1.00}
& 62.01\dn{5.53}
& 96.18
\\

\rowcolor{TableStripe}
PACE~\citep{wei2026pace}
& 70.38\dn{2.55}
& 55.75\dn{25.89}
& 43.53\dn{1.05}
& 64.02\dn{6.98}
& 58.42\dn{9.12}
& 44.12
\\

SelfCompact~\citep{li2026self}
& 76.83\up{3.90}
& 79.62\dn{2.02}
& 36.20\dn{8.38}
& 71.00
& 65.91\dn{1.63}
& 62.70
\\

\rowcolor{TableStripe}
ACON-Core~\citep{kang2025acon}
& 71.79\dn{1.14}
& 68.55\dn{13.09}
& \second{47.95}\up{3.37}
& 65.00\dn{6.00}
& 63.32\dn{4.22}
& 56.13
\\

Self-GC~\citep{hao2026self}
& 63.45\dn{9.48}
& 71.67\dn{9.97}
& 43.71\dn{0.87}
& 71.00
& 62.46\dn{5.08}
& 66.21
\\

\rowcolor{TableStripe}
LRE~\citep{jahan2026learning}
& 62.62\dn{10.31}
& 63.44\dn{18.20}
& 27.14\dn{17.44}
& 68.00\dn{3.00}
& 55.30\dn{12.24}
& 91.64
\\

CoMem~\citep{zhang2026comem}
& 56.43\dn{16.50}
& 59.86\dn{21.78}
& 2.50\dn{42.08}
& 58.00\dn{13.00}
& 44.20\dn{23.34}
& \best{30.64}
\\

\rowcolor{TableStripe}
SAM~\citep{hu2026sam}
& 67.62\dn{5.31}
& 81.63\dn{0.01}
& 47.24\up{2.66}
& 67.00\dn{4.00}
& 65.87\dn{1.67}
& 53.85
\\

SWE-Pruner~\citep{wang2026swe}
& 63.45\dn{9.48}
& 72.60\dn{9.04}
& 40.87\dn{3.71}
& 65.00\dn{6.00}
& 60.48\dn{7.06}
& 60.07
\\

\rowcolor{TableStripe}
Sculptor~\citep{li2026sculptor}
& 51.19\dn{21.74}
& 81.62\dn{0.02}
& 35.25\dn{9.33}
& 69.00\dn{2.00}
& 59.26\dn{8.28}
& 51.75
\\

\hline

ACM~\citep{li2026acm}
& 38.69\dn{34.24}
& 55.93\dn{25.71}
& 6.20\dn{38.38}
& 34.00\dn{37.00}
& 33.70\dn{33.84}
& 94.56
\\

\hline

\rowcolor{TableOurs}
\textbf{\method{} (Ours)}
& 76.03\up{3.10}
& \second{84.12}\up{2.48}
& \best{51.01}\up{6.43}
& 75.00\up{4.00}
& \best{71.54}\up{4.00}
& \second{43.76}
\\

\Xhline{1.2pt}

\end{tabular}
}
\vspace{-10pt}
\end{table*}

On Eval40, the reported compression run obtains reward $0.7154$ with 43.76M
tokens, compared with $0.6754$ and 75.50M for the baseline. This corresponds to $42.04\%$ fewer tokens.
The token total includes 0.077M summary tokens.

Sliding windows with $K=5,10,20$ obtain rewards of $0.5639$, $0.6212$, and
$0.6867$. The largest window still uses 80.82M tokens. Periodic summarization
with $n=5$ reaches $0.7017$ reward but uses 94.85M tokens, including 3.01M
summary tokens. PACE uses 44.12M tokens with reward $0.5842$. Within this comparison group, \method{} achieves the highest mean reward with 43.76M
total tokens. This supports evaluating compression timing together with
the cost of generating summaries, rather than reporting the length of a
shortened prompt alone. The comparison also illustrates that reducing visible history alone does not
guarantee lower total consumption, since altered execution paths and repeated
summary operations can offset prompt-level savings.

Table~\ref{tab:method_comparison} also reports additional published
context-management baselines evaluated on the same fixed Eval40 tasks.

\subsection{Cross-Model Evaluation}
\label{sec:cross_model}

Table~\ref{tab:cross_model} applies the same compression procedure to four
acting models on the same fixed Eval40 tasks. GLM-5.3-Flash
\citep{zeng2026glm} improves from $0.6170$ to $0.6492$ reward while tokens
fall from 47.58M to 38.49M. MiMo-V2.5 \citep{mimov25} changes from $0.5196$
to $0.5252$ with tokens falling from 53.55M to 18.91M. Both runs improve the
observed quality and token metrics together.

Qwen3.8-Flash \citep{qwen3.8flashnext} improves reward from $0.7249$ to
$0.7784$, while tokens fall from 49.75M to 38.23M. Hunyuan reduces tokens from 41.00M to 12.83M while reward increases from
$0.6019$ to $0.6391$. Token usage decreases for all four reported backbones, while the magnitude
and domain distribution of reward changes remain model dependent.
These results show that the procedure transfers across different acting
models, although the magnitude of the efficiency gain and the associated
reward changes remain backbone dependent. In particular, token totals alone
do not identify whether a change is caused by longer execution, additional
retries, or larger requests. Domain scores are given in
Appendix~\ref{app:online_results}.

\begin{table*}[t]
\centering
\caption{
\textbf{Cross-model evaluation on Eval40.}
Scores are reward multiplied by 100.
Colored arrows show changes relative to the same backbone without StateComp.
Tokens report the total model usage in millions. Each domain contains
ten tasks, so Avg. is the mean of the four domain scores.
}
\vspace{+3pt}
\label{tab:cross_model}

\small
\renewcommand{\arraystretch}{1.08}
\setlength{\tabcolsep}{4.6pt}

\resizebox{\textwidth}{!}{%
\begin{tabular}{ll|cccccc}

\Xhline{1.2pt}

\rowcolor{TableHeader}
\textbf{Backbone}
&
\textbf{Method}
&
\textbf{Code}
&
\textbf{Office}
&
\textbf{Sec.}
&
\textbf{Web}
&
\textbf{Avg.}
&
\textbf{Tokens (M)}
\\

\Xhline{1.0pt}

% ==================== Hunyuan ====================

&
Base
&
55.30
&
79.24
&
\textbf{60.24}
&
46.00
&
60.19
&
41.00
\\

\rowcolor{TableOurs}
\cellcolor{white}\multirow{-2}{*}{Hunyuan}
&
\textbf{\method{}}
&
\textbf{64.31}\up{9.01}
&
\textbf{82.20}\up{2.96}
&
60.14\dn{0.10}
&
\textbf{49.00}\up{3.00}
&
\textbf{63.91}\up{3.72}
&
\textbf{12.83}
\\

\hline

% ==================== Qwen ====================

&
Base
&
75.25
&
83.42
&
45.28
&
86.00
&
72.49
&
49.75
\\

\rowcolor{TableOurs}
\cellcolor{white}\multirow{-2}{*}{Qwen3.8-Flash}
&
\textbf{\method{}}
&
\textbf{93.46}\up{18.21}
&
\textbf{88.71}\up{5.29}
&
41.20\dn{4.08}
&
\textbf{88.00}\up{2.00}
&
\textbf{77.84}\up{5.35}
&
38.23
\\

\hline

% ==================== GLM ====================

&
Base
&
79.49
&
66.13
&
34.67
&
66.51
&
61.70
&
47.58
\\

\rowcolor{TableOurs}
\cellcolor{white}\multirow{-2}{*}{GLM-5.3-Flash}
&
\textbf{\method{}}
&
66.15\dn{13.34}
&
69.90\up{3.77}
&
38.64\up{3.97}
&
85.00\up{18.49}
&
64.92\up{3.22}
&
38.49
\\

\hline

% ==================== MiMo ====================

&
Base
&
53.76
&
77.05
&
22.04
&
55.00
&
51.96
&
53.55
\\

\rowcolor{TableOurs}
\cellcolor{white}\multirow{-2}{*}{MiMo-V2.5}
&
\textbf{\method{}}
&
57.65\up{3.89}
&
71.64\dn{5.41}
&
22.77\up{0.73}
&
58.00\up{3.00}
&
52.52\up{0.56}
&
18.91
\\

\Xhline{1.2pt}

\end{tabular}
}
\vspace{-15pt}
\end{table*}

\section{Conclusion}
\label{sec:conclusion}

We introduced \method{}, which predicts whether past interactions still need
their full details at the current agent state. Two stage annotation supplies
supported compression boundaries, hidden states provide prediction features,
and span and token gates determine when selected history is summarized.
Committed summaries become part of the effective history without restoring
removed raw interactions. The reported full benchmark comparison reduces
agent and summary tokens by approximately half while maintaining average
reward. These findings support separating the prediction of compression opportunities
from the decision to rewrite a sufficiently large span. The experiments measure complete-task token usage; they do not
isolate the effect of the span gates on cache reuse.

% Required by the ICLR 2027 author policy. Authors should check that this
% statement describes all AI use in the final study and submission form.
\subsection*{AI Use Statement}
Generative AI tools were used for language editing and literature retrieval and discovery, including identifying potentially relevant papers, search keywords, and related work, as well as surveying the current state of research on relevant topics. An LLM was also used in the annotation procedure to assist in constructing the compression supervision described in the paper and appendix. The annotation procedure and resulting labels were reviewed by the authors. The final selection, interpretation, and presentation of the literature and experimental results were determined by the authors. All AI-assisted content was reviewed by the authors, who take full responsibility for the final content of this work.

\subsection*{Reproducibility Statement}

Appendices A--E describe the two-stage annotation procedure, grouped
training and evaluation splits, representation construction, router
objectives, theoretical analysis, and online compression algorithm.
Appendices F and G provide the fixed benchmark protocols, supplementary
results, operating settings, and the fixed Eval40 task identifiers.
Token-accounting conventions and the compression configuration are
specified alongside the corresponding experiments.

\clearpage
\setlength{\bibsep}{.5ex plus .8ex}
\IfFileExists{iclr2027_conference.bib}{%
  \IfFileExists{iclr2027_conference.bst}{%
    \bibliographystyle{unsrtnat}
  }{%
    \bibliographystyle{unsrtnat}
  }
  \bibliography{iclr2027_conference}
}{%
  \PackageWarningNoLine{StateComp}{The existing iclr2027_conference.bib file is required to resolve citations}
}
\clearpage
\clearpage
\appendix
\setcounter{figure}{0}
\setcounter{table}{0}
\setcounter{equation}{0}
\renewcommand{\thefigure}{A\arabic{figure}}
\renewcommand{\thetable}{A\arabic{table}}
\numberwithin{equation}{section}
\renewcommand{\theHequation}{statecomp.appendix.\thesection.\arabic{equation}}
\renewcommand{\theHfigure}{statecomp.appendix.\arabic{figure}}
\renewcommand{\theHtable}{statecomp.appendix.\arabic{table}}
\providecommand{\method}{\textsc{StateComp}}
\providecommand{\na}{\textnormal{n/a}}
\providecolor{TableHeader}{RGB}{226,224,234}
\providecolor{TableStripe}{RGB}{246,246,246}
\providecolor{TableOurs}{RGB}{237,236,247}
\providecommand{\apptabsetup}{%
  \small\setlength{\tabcolsep}{4pt}%
  \renewcommand{\arraystretch}{1.12}\arrayrulecolor{black}}
\newsavebox{\scfinaltablebox}
\providecommand{\appfitbox}[1]{%
  \sbox{\scfinaltablebox}{#1}%
  \ifdim\wd\scfinaltablebox>\linewidth
    \resizebox{\linewidth}{!}{\usebox{\scfinaltablebox}}%
  \else\usebox{\scfinaltablebox}\fi}
\providecommand{\appfigure}[2][width=0.90\linewidth]{\includegraphics[#1]{#2}}
\newcounter{scfinalprop}[section]
\renewcommand{\thescfinalprop}{\thesection.\arabic{scfinalprop}}
\newenvironment{scproposition}[1]{%
  \refstepcounter{scfinalprop}\par\medskip\noindent
  \textbf{Proposition \thescfinalprop\ (#1).}\ \itshape}{\par\medskip}
\newenvironment{scproof}{\par\noindent\textit{Proof.}\ }{\hfill$\square$\par\medskip}
\setlength{\emergencystretch}{2em}
\renewcommand{\topfraction}{0.95}
\renewcommand{\textfraction}{0.05}
\renewcommand{\floatpagefraction}{0.8}

\section{Compression Targets and Supervision}
\label{app:experimental_details}

This appendix provides the supervision, training details, mathematical
analysis, and supplementary experiments for \method{}. The analytical
results concern the stated prediction and execution rules; empirical
compression quality is evaluated separately.

\subsection{Interactions and Effective History}

An interaction $S_i=(R_i,A_i,O_i)$ contains reasoning, an assistant action,
and its observation. Tool calls and their responses remain within complete
interaction boundaries. Before step $k$, let $P_k=(S_1,\ldots,S_{k-1})$
denote the original prefix and $H_k$ the effective history after previously
committed replacements. Let $\mathcal I_k$ index the surviving raw
interactions. The agent and router use $H_k$; stored original trajectories
are not available as an online retrieval channel.

The target is defined jointly by a historical interaction and the current
state:
\begin{equation}
 y_{i,k}=\begin{cases}
 0\quad(\mathrm{KEEP}), & \text{retain the original details},\\
 1\quad(\mathrm{READY}), & \text{replace the details by an accurate summary}.
 \end{cases}
 \label{eq:app_pair_target}
\end{equation}
READY does not mean that every fact is irrelevant. Facts still needed by
the task must remain in the summary or elsewhere in the effective history.
The online router predicts
\begin{equation}
 p_{i,k}=p_\theta(S_i,H_k),\qquad
 \widehat y_{i,k}=\mathbf 1\{p_{i,k}\ge\tau\},\qquad i\in\mathcal I_k.
 \label{eq:app_pair_decision}
\end{equation}
Selected interactions may occupy several disconnected regions; a recent
window is not imposed on candidate eligibility.

\begin{table}[htbp]
\centering
\caption{\textbf{Notation for supervision and online compression.}}
\label{tab:app_notation}
\apptabsetup
\begin{tabular}{l|p{0.68\linewidth}}
\Xhline{1.1pt}
\rowcolor{TableHeader}
\textbf{Symbol} & \textbf{Meaning}\\
\Xhline{0.8pt}
$S_i$ & Complete historical interaction\\
\rowcolor{TableStripe}
$P_k$ & Uncompressed reference prefix before interaction $k$\\
$H_k,\mathcal I_k$ & Effective history and identifiers of surviving raw interactions\\
\rowcolor{TableStripe}
$y_{i,k}$ & Reference label: KEEP $=0$, READY $=1$\\
$t_i^*$ & Earliest supported compression checkpoint\\
\rowcolor{TableStripe}
$F_\phi,g_\theta$ & Frozen representation model and trained router\\
$h_i,q_k,r_{i,k}$ & Historical, current, and aggregated historical features\\
\rowcolor{TableStripe}
$z_{i,k},p_{i,k}$ & Router logit and sigmoid score\\
$\tau$ & Router selection threshold\\
\rowcolor{TableStripe}
$K_{\mathrm{ctx}},K_{\mathrm{sw}},K_{\mathrm{hist}}$ & Recent context size, sliding window size, and same-prefix feature selection count\\
$B,\kappa,B_{\min}$ & Continuous source span, interaction threshold, and source token threshold\\
\rowcolor{TableStripe}
$M,\operatorname{Tok}(H)$ & Per-view representation budget and serialized history length\\
\Xhline{1.1pt}
\end{tabular}
\end{table}

\subsection{Two-Stage Annotation}

Stage 1 independently annotates each checkpoint and produces a
structured judgment for every historical interaction at that checkpoint.
Each record includes task progress, observation complexity, expected
requirements of the next subtask, the remaining utility of the interaction,
a KEEP/READY judgment, and the locations of supporting evidence. Positive
evidence must already occur in $P_k$. Future interactions can reveal a
missing dependency and veto a proposed replacement, but cannot justify an
earlier replacement using information that was not yet available.

Writing $e_{i,k}$ for supporting prefix evidence and $v_{i,k}$ for a future
dependency veto, a positive local judgment $b_{i,k}$ obeys
\begin{equation}
 b_{i,k}=1\ \Longrightarrow\ e_{i,k}=1\ \text{and}\ v_{i,k}=0.
 \label{eq:app_evidence_rule}
\end{equation}
Stage 2 reviews these records across checkpoints for the same interaction,
checks proposed boundaries against their supporting evidence, and selects
the earliest supported compression point $t_i^*$. Unresolved cases remain
KEEP. The final labels are
\begin{equation}
 y_{i,k}=\mathbf 1\{k\ge t_i^*\},\qquad
 t_i^*\in\{i+1,\ldots,T\}\cup\{\infty\}.
 \label{eq:app_boundary_encoding}
\end{equation}
The value $\infty$ denotes an interaction with no supported boundary.
Monotonicity is part of this encoding for the annotated continuation, not
a guarantee for every possible future trajectory. Online predictions are
therefore recomputed after the effective history changes.

\subsection{Data and Grouped Evaluation}

The supervision collection contains 300 trajectories and 244,526
interaction--state pairs. READY accounts for 15,633 pairs, giving
\begin{equation}
 \pi=\frac{15{,}633}{244{,}526}\approx0.06393.
 \label{eq:app_prevalence}
\end{equation}
An always-KEEP classifier consequently reaches approximately $93.61\%$
accuracy without identifying a single compression opportunity. We use
ranking and precision--recall metrics rather than accuracy alone.

\begin{table}[htbp]
\centering
\caption{\textbf{Compression supervision and evaluation partitions.}}
\label{tab:app_data_statistics}
\apptabsetup
\begin{tabular}{l|r}
\toprule
\rowcolor{TableHeader}\textbf{Quantity} & \textbf{Value}\\
\midrule
Trajectories & 300\\
Recorded states & 7,931\\
Checkpoint groups & 7,631\\
Interaction--state pairs & 244,526\\
KEEP pairs & 228,893\\
READY pairs & 15,633\\
Supported boundaries & 1,832\\
Interactions without a supported boundary & 6,099\\
\midrule
Training / validation / test trajectories per rotation & 200 / 50 / 50\\
Grouped folds & 5\\
\bottomrule
\end{tabular}
\end{table}

All pairs from the same trajectory remain in one partition. Model
configuration, training checkpoint, class weighting, and prediction
threshold are selected using training and validation data. The five
rotations evaluate generalization across trajectories rather than across
randomly separated pairs from the same trajectory.

\section{State Representations and Router Learning}
\label{app:router_protocol}

\subsection{Historical and Current-State Features}

The grouped router study uses a frozen Qwen2.5-7B representation model.
Let $X_k=\operatorname{Serialize}(P_k)$, and let $\ell_i$ be the last token
of $S_i$. A full-context pass of the frozen representation model $F_\phi$
provides
\begin{equation}
 h_i=F_\phi(X_k)_{\ell_i},\qquad q_k=F_\phi(X_k)_{|X_k|}.
 \label{eq:app_full_features}
\end{equation}
The first vector describes the historical interaction; the second describes
the current state. A notation covering the concatenation-based variants is
\begin{equation}
 r_{i,k}=\operatorname{Agg}\{h_j:j\in\mathcal J_{i,k}\},\qquad
 x_{i,k}=[h_i;q_k;r_{i,k}],\qquad p_{i,k}=\sigma(g_\theta(x_{i,k})).
 \label{eq:app_router_features}
\end{equation}
Here $\mathcal J_{i,k}$ lies within the available prefix. The experiments
compare features with and without recurrent aggregation and same-prefix
historical selection. These operations construct prediction features;
they do not restore removed raw interactions.

\begin{scproposition}{Causal prefix invariance}
\label{prop:app_prefix}
Consider a deterministic causal Transformer with fixed parameters,
unchanged prefix tokens and positional assignments, and no transformation
of the prefix that depends on the length of an appended suffix. In exact
arithmetic, appending tokens leaves every existing prefix hidden state
unchanged.
\end{scproposition}
\begin{scproof}
Let $H_j^{(\ell)}$ be the representation at prefix position $j$ and layer
$\ell$. The input states $H_j^{(0)}$ are identical. For one attention head,
\begin{equation}
 A_j^{(\ell)}=\sum_{u\le j}\alpha_{ju}^{(\ell)}W_V^{(\ell)}H_u^{(\ell)},
 \qquad
 \alpha_{ju}^{(\ell)}=
 \frac{\exp(a_{ju}^{(\ell)})}{\sum_{v\le j}\exp(a_{jv}^{(\ell)})}.
 \label{eq:sc_causal_attention}
\end{equation}
Assume that the states at all prefix positions are unchanged at layer
$\ell$. Each query, permitted key, value, and positional contribution in
Equation~\eqref{eq:sc_causal_attention} is then unchanged. The normalization
contains no suffix positions, so every attention output is unchanged.
Residual connections, tokenwise normalization, and feed-forward layers
preserve the equality. Induction over layers proves the result.
\end{scproof}

Thus, a historical vector cannot become state dependent merely because
new tokens are appended. The router obtains state dependence from $q_k$
or from a freshly constructed input containing both the target interaction
and current context. The proposition also characterizes reusable prefix
states after an online rewrite.

\subsection{Weighted Classification}

With READY encoded as one, the router objective is
\begin{equation}
 \mathcal L(\theta)=-\frac1N\sum_{n=1}^N
 \left[w_1y_n\log p_n+w_0(1-y_n)\log(1-p_n)\right].
 \label{eq:app_weighted_bce}
\end{equation}
The weighted BCE experiments set $w_0=1$ and vary $w_1$. The representation
model remains frozen. For $p=\sigma(z)$ and a binary label $y$,
\begin{align}
 \frac{\partial\ell}{\partial z}
 &=\bigl(w_1y+w_0(1-y)\bigr)(p-y),\label{eq:app_loss_gradient}\\
 \frac{\partial^2\ell}{\partial z^2}
 &=\bigl(w_1y+w_0(1-y)\bigr)p(1-p).
 \label{eq:sc_loss_curvature}
\end{align}
These derivatives describe the loss as a function of a logit, not convexity
in the parameters of a nonlinear router.

\begin{scproposition}{Weighted score and underlying posterior}
\label{prop:app_weighted_posterior}
Let $\eta(x)=\Pr(Y=1\mid x)$ and $w_0,w_1>0$. For $0<\eta(x)<1$, the
unique minimizer of the conditional weighted BCE is
\begin{equation}
 p^*(x)=\frac{w_1\eta(x)}{w_1\eta(x)+w_0(1-\eta(x))},\qquad
 \operatorname{logit}p^*=
 \operatorname{logit}\eta+\log\frac{w_1}{w_0}.
 \label{eq:app_weighted_optimum}
\end{equation}
\end{scproposition}
\begin{scproof}
The conditional risk is $R(p)=-w_1\eta\log p-w_0(1-\eta)\log(1-p)$.
Its derivative vanishes when
$w_1\eta(1-p)=w_0(1-\eta)p$, giving the first expression. Moreover,
\begin{equation}
 R''(p)=\frac{w_1\eta}{p^2}+\frac{w_0(1-\eta)}{(1-p)^2}>0,
\end{equation}
which proves uniqueness. Taking the ratio $p^*/(1-p^*)$ proves the
log-odds identity. Boundary cases follow by continuity. If
$D=w_1\eta+w_0(1-\eta)$, the excess risk also satisfies
\begin{equation}
 R(p)-R(p^*)=D\,\operatorname{KL}
 \bigl(\operatorname{Bern}(p^*)\,\|\,\operatorname{Bern}(p)\bigr),
 \label{eq:sc_weighted_excess}
\end{equation}
obtained by collecting the two logarithmic terms.
\end{scproof}

The corresponding inverse transformation is
\begin{equation}
 \eta=\frac{w_0p^*}{w_1(1-p^*)+w_0p^*}.
 \label{eq:app_weighted_inverse}
\end{equation}
This population identity explains why class weighting changes score
interpretation. It does not establish calibration of the fitted router.

\subsection{Asymmetric Costs and Operating Thresholds}

Let $C_{\mathrm{unsafe}}$ denote the cost of selecting a KEEP interaction
and $C_{\mathrm{miss}}$ the cost of retaining a READY interaction. Both
are positive. Correct label decisions have zero cost in this analysis;
summary-generation errors are considered separately in
Appendix~\ref{app:online_algorithm}.

\begin{scproposition}{Cost-sensitive selection}
\label{prop:app_bayes}
If $\eta(x)$ is known, the minimum conditional label risk is attained by
selection whenever
\begin{equation}
 \eta(x)\ge\eta_0:=\frac{C_{\mathrm{unsafe}}}
 {C_{\mathrm{unsafe}}+C_{\mathrm{miss}}}.
 \label{eq:app_bayes_threshold}
\end{equation}
For the ideal weighted score, the equivalent threshold is
\begin{equation}
 \tau_w=\frac{w_1C_{\mathrm{unsafe}}}
 {w_1C_{\mathrm{unsafe}}+w_0C_{\mathrm{miss}}}.
 \label{eq:app_weighted_threshold}
\end{equation}
\end{scproposition}
\begin{scproof}
The conditional costs of selection and retention are
$C_{\mathrm{unsafe}}(1-\eta)$ and $C_{\mathrm{miss}}\eta$, respectively.
Comparing them gives Equation~\eqref{eq:app_bayes_threshold}.
The map $\eta\mapsto p^*$ in
Equation~\eqref{eq:app_weighted_optimum} is strictly increasing.
Substitution of $\eta_0$ therefore gives the equivalent weighted threshold.
More explicitly, a decision that disagrees with this cost-optimal decision
has excess conditional risk
\begin{equation}
 (C_{\mathrm{unsafe}}+C_{\mathrm{miss}})|\eta-\eta_0|.
 \label{eq:sc_cost_regret}
\end{equation}
This follows by subtracting the smaller of the two conditional costs from
the larger one.
\end{scproof}

Class weighting and conservative decision thresholds have different roles.
The former changes the training loss; the latter determines the operating
tradeoff. In the experiments, operating choices use validation data rather
than a known posterior or an assumed numerical cost ratio.

\section{Bounded Representation and Efficiency}
\label{app:representation_design}

\subsection{Input Construction and Computational Cost}

For each surviving historical target, the bounded construction forms
\begin{equation}
 V_{i,k}=\mathcal B_M(S_i,H_k;K_{\mathrm{ctx}}),\qquad
 |V_{i,k}|\le M,\qquad
 \widetilde h_{i,k}=F_\phi(V_{i,k})_{|V_{i,k}|}.
 \label{eq:app_bounded_features}
\end{equation}
It combines the target with current context and a limited portion of the
effective history. The representation comparison uses a 4,096-token base
budget and up to 1,024 additional target tokens, for $M=5,120$. Recent4
specifies a context construction, not the minimum span length for an online
compression operation.

Let $L_k$ denote full-context length, $N_k$ the number of surviving targets,
$\mathcal F(n)$ the cost of a length-$n$ model pass, and $c_g$ the router
cost per target. Fresh full-context extraction and target-specific bounded
extraction have workloads
\begin{equation}
 C_{\mathrm{full}}(k)=\mathcal F(L_k)+O(N_kc_g),\qquad
 C_{\mathrm{bounded}}(k)\le N_k\mathcal F(M)+O(N_kc_g).
 \label{eq:app_extraction_cost}
\end{equation}
One full pass can supply all historical readouts. Conversely, bounded
extraction uses multiple short views, which may be batched; its total work
still depends on $N_k$. For a dense-attention model with $L$ layers and
width $d$, the usual arithmetic model is
$\mathcal F(n)=O(L(nd^2+n^2d))$. Across a trajectory,
\begin{equation}
 C_{\mathrm{trajectory}}\le
 \sum_k N_k\mathcal F(M)+\sum_k O(N_kc_g).
 \label{eq:app_trajectory_cost}
\end{equation}
A bounded view therefore controls individual input size without implying
constant cost per checkpoint.

\subsection{Stability of Decisions and Selected Spans}

\begin{scproposition}{Bounded perturbations preserve routing and spans}
\label{prop:app_margin}
Suppose full and bounded features $x_i,\widetilde x_i$ are supplied to the
same logit function $g$, with
$|g(x_i)-g(\widetilde x_i)|\le L_g\epsilon_i$. Let
$a_\tau=\log(\tau/(1-\tau))$, $0<\tau<1$. The decision for target $i$
is unchanged if
\begin{equation}
 |g(x_i)-a_\tau|>L_g\epsilon_i.
 \label{eq:app_margin}
\end{equation}
If this holds for every eligible interaction, and the source history and
execution gates are unchanged, the two feature constructions produce the
same maximal selected spans and the same gate-eligible candidates.
\end{scproposition}
\begin{scproof}
A score more than $L_g\epsilon_i$ above the threshold remains above it
after the perturbation; a score more than that amount below it remains
below it. Since the sigmoid is strictly increasing, thresholding its
output is equivalent to thresholding the logit at $a_\tau$. This proves
pointwise decision equality. Equality for every target gives identical
binary masks. Maximal runs are determined uniquely by the mask and the
fixed boundaries of eligible interactions. Each resulting run consequently
has the same source length and token count, so both execution gates return
the same answer.
\end{scproof}

The result concerns a fixed router under a stated perturbation bound.
The empirical ablations below compare separately fitted feature variants;
they measure performance rather than establish the bound in
Equation~\eqref{eq:app_margin}.

\subsection{Representation Measurements}

The input-size study covers 110 trajectories and 2,558 states. Mean bounded
input is 4,181.5 tokens; 86 trajectories exceed the base budget, 73 reach
the 5,120-token limit, and 178 current-step inputs are truncated. Timing
uses seven states on which both implementations complete on the same
device. The representation-quality comparison is separate from the
boundary-label router study in Appendix~\ref{app:router_ablation}.

\begin{table}[htbp]
\centering
\caption{\textbf{Representation efficiency.} Input statistics, paired runtime, and classification quality refer to their respective measurements.}
\label{tab:app_efficiency_full}
\apptabsetup
\appfitbox{%
\begin{tabular}{l|rr}
\Xhline{1.1pt}
\rowcolor{TableHeader}
\textbf{Metric} & \textbf{Full history} & \textbf{Bounded}\\
\Xhline{0.8pt}
Mean input tokens & \na & 4,181.50\\
\rowcolor{TableStripe}
Largest recorded input & 203,675 & 5,120\\
Representative long input & 127,833 & 5,120\\
\rowcolor{TableStripe}
Paired extraction time (s) & 76.40 & 6.03\\
Mean latency (s/state) & 10.914 & 0.861\\
\rowcolor{TableStripe}
Peak GPU memory (GiB) & 29.426 & 18.620\\
Forward workspace (GiB) & 11.867 & 1.062\\
\rowcolor{TableStripe}
Throughput (states/s) & 0.0916 & 1.1611\\
Separate test F1 & 0.7000 & 0.7818\\
\Xhline{1.1pt}
\end{tabular}}
\end{table}

Total extraction time decreases from 76.40 to 6.03 seconds, a
$12.67\times$ measured speedup. Peak GPU allocation and forward workspace
measure different quantities. The full-model memory result comes from a
successful 57,401-token input; inputs of 131,576 and 203,675 tokens exhaust
GPU memory and are not included in the paired timing. This speedup is for
representation extraction, not complete-task execution.

\begin{table}[htbp]
\centering
\caption{\textbf{Context and historical-feature ablations in the representation comparison.}}
\label{tab:app_context_ablation}
\apptabsetup
\appfitbox{%
\begin{tabular}{l|rr}
\Xhline{1.1pt}
\rowcolor{TableHeader}
\textbf{Representation} & \textbf{Test F1} & \textbf{Balanced accuracy}\\
\Xhline{0.8pt}
Full Context & 0.7000 & 0.5951\\
\rowcolor{TableStripe}
Recent2 & 0.7101 & 0.5940\\
Recent4 & 0.7340 & 0.6036\\
\rowcolor{TableStripe}
Local + GRU & 0.7531 & 0.5979\\
Local + same-prefix Top $k$ & 0.7545 & 0.6022\\
\rowcolor{TableStripe}
Hybrid & 0.6749 & 0.6011\\
Capped Recent4 & 0.7518 & 0.5867\\
\rowcolor{TableStripe}
\quad + GRU & 0.7465 & 0.6048\\
\quad + same-prefix Top $k$ & 0.7477 & \textbf{0.6156}\\
\hline\rowcolor{TableOurs}
\quad + Hybrid & \textbf{0.7818} & 0.5988\\
\Xhline{1.1pt}
\end{tabular}}
\end{table}

The capped hybrid achieves the highest F1, while capped same-prefix
selection achieves the highest balanced accuracy. The full set of rows
shows that adding recurrent or historical features is not uniformly
beneficial. These are comparisons of particular constructions, not a claim
that bounded inputs preserve all information in the full history.

\section{Compression Dynamics and Prediction Diagnostics}
\label{app:compression_dynamics}
\label{app:router_ablation}

\subsection{Readiness and Compression Delay}

At checkpoint $k$, the annotated readiness fraction is
\begin{equation}
 a_k=\frac1{k-1}\sum_{i<k}y_{i,k}.
 \label{eq:app_readiness}
\end{equation}
Even under monotone boundary labels, this fraction need not increase at
every step. If $u_k$ old interactions newly become READY and the new
interaction has label $b_k=y_{k,k+1}$, then
\begin{equation}
 a_{k+1}-a_k=\frac{u_k+b_k-a_k}{k}.
 \label{eq:sc_readiness_increment}
\end{equation}
Indeed, the previous READY count is $(k-1)a_k$, and the next count is
$(k-1)a_k+u_k+b_k$. Dividing by the new denominator $k$ and subtracting
$a_k$ proves the identity. New KEEP interactions can lower the fraction
without reversing any existing boundary label.

\begin{figure}[htbp]
\centering
\appfigure[width=.48\linewidth]{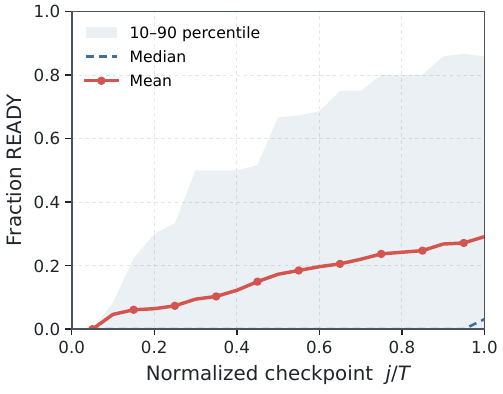}
\caption{\textbf{Variation in compression opportunities.} Mean and median READY fractions and the 10th--90th percentile range across trajectories.}
\label{fig:app_overall_readiness}
\end{figure}

The mean, median, and percentile range summarize variation across
trajectories. The percentile band describes that variation rather than
uncertainty in an estimated mean. Compression delay additionally describes
how long an interaction remains in its original form before a supported
boundary. For normalized delays $d_i$ in a finite-boundary group $G$, its
empirical distribution is
\begin{equation}
 \widehat F_G(u)=|G|^{-1}\sum_{i\in G}\mathbf 1\{d_i\le u\}.
 \label{eq:app_delay_cdf}
\end{equation}
This distribution is conditional on a finite supported boundary.

\begin{figure}[htbp]
\centering
\appfigure[width=.92\linewidth]{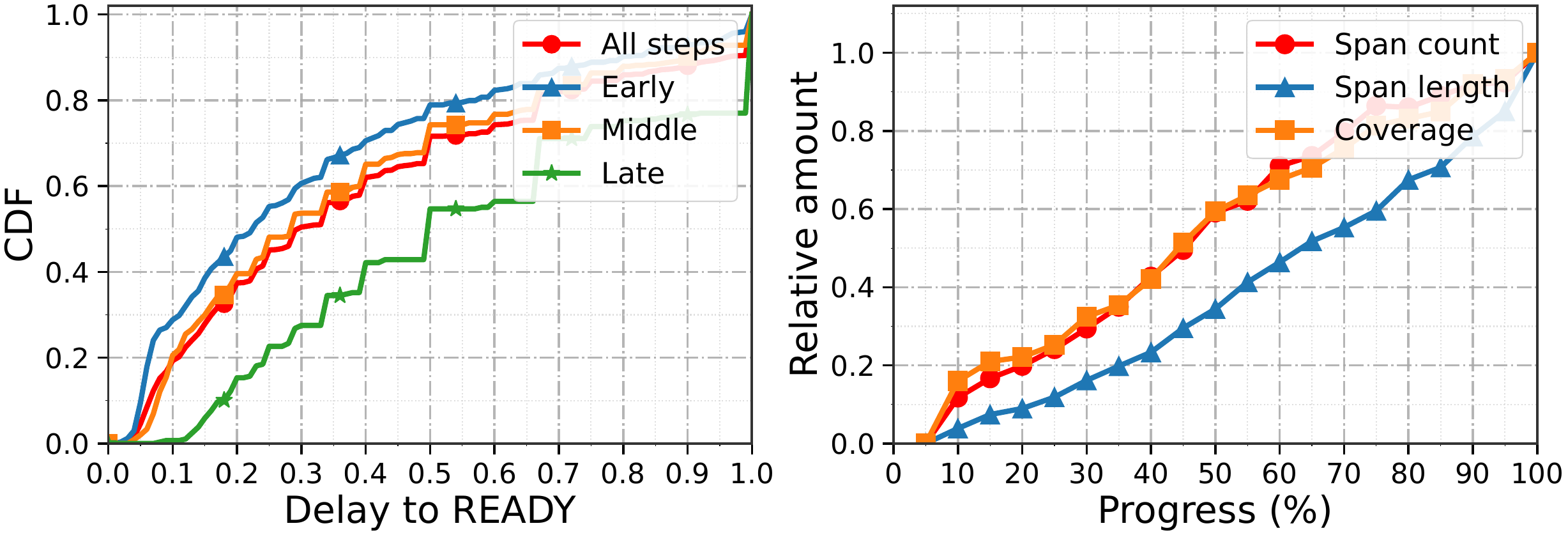}
\caption{\textbf{Compression delay and span dynamics.} Left: normalized delay to READY by interaction position. Right: relative span count, span length, and coverage over task progress.}
\label{fig:span_formation}
\end{figure}

For a binary selection mask $a_1,\ldots,a_N$ within a consecutive
eligible region, let $a_0=0$. Then
\begin{equation}
 J=\sum_{i=1}^Na_i(1-a_{i-1}),\qquad
 c=N^{-1}\sum_{i=1}^Na_i,\qquad
 \overline\ell=\frac{\sum_i a_i}{J}\quad(J>0).
 \label{eq:app_span_statistics}
\end{equation}
Each maximal selected run contributes one transition from zero to one,
which proves the expression for $J$. Protected messages and existing
summaries separate eligible regions. Thus, coverage, number of spans,
and average span length describe different aspects of the selected history.

\subsection{Router Configurations}
\label{app:hidden_geometry}

\begin{table}[htbp]
\centering
\caption{\textbf{Router configurations and operating points.} PR AUC and AUROC are ranking metrics; precision and recall are percentages at the reported operating point. \textnormal{n/a} denotes an unreported metric.}
\label{tab:app_router_summary}
\apptabsetup
\appfitbox{%
\begin{tabular}{l|rrrr}
\Xhline{1.1pt}
\rowcolor{TableHeader}
\textbf{Configuration} & \textbf{PR AUC} & \textbf{AUROC} & \textbf{Precision} & \textbf{Recall}\\
\Xhline{0.8pt}
BCE baseline & 0.1685 & 0.6870 & \na & \na\\
\rowcolor{TableStripe}
Weighted BCE ($w_1=10$) & 0.1736 & 0.7096 & \na & \na\\
Random oversampling 1:4 & 0.1674 & \na & \na & \na\\
\rowcolor{TableStripe}
Linear & 0.1425 & \na & \na & \na\\
MLP 128 & 0.1455 & \na & \na & \na\\
\rowcolor{TableStripe}
MLP 2048 & 0.1655 & \na & \na & \na\\
Candidate + current + Top4 & 0.1455 & 0.7171 & 28.571 & 0.081\\
\rowcolor{TableStripe}
CB Focal ($\beta=0.9999,\gamma=3$) & 0.1624 & \na & \na & \na\\
Weighted BCE ($w_1=5$) & \na & \na & 72.222 & 0.111\\
\rowcolor{TableStripe}
Positive ratio sweep & 0.1669 & \na & 58.970 & \na\\
100\% training trajectories & \na & \na & 62.319 & 0.092\\
\Xhline{1.1pt}
\end{tabular}}
\end{table}

Weighted BCE with $w_1=10$ reaches AUROC 0.7096 and PR AUC 0.1736.
The separate $w_1=5$ operating point has precision $72.222\%$ at recall
$0.111\%$, illustrating the limited coverage of a highly selective rule.
The ranking result and high-precision result are not attributed to a
single fitted configuration. These operating-point measurements are not
a guarantee of safe online compression.

\begin{figure}[htbp]
\centering
\appfigure[width=.90\linewidth]{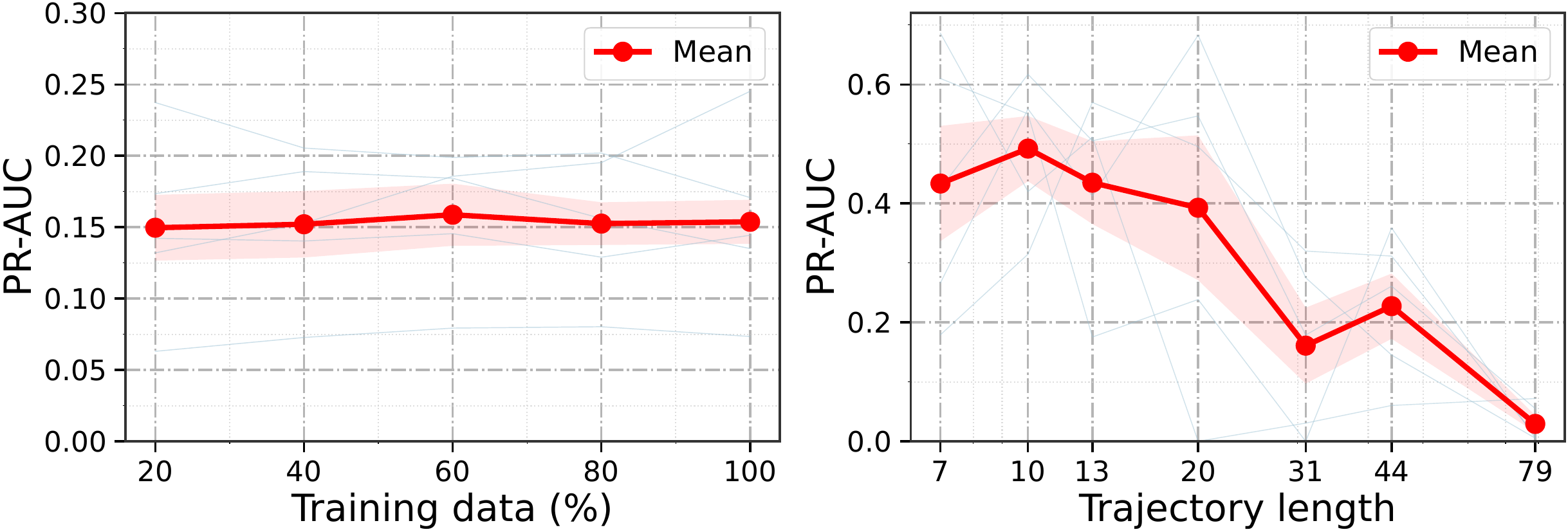}
\caption{\textbf{Prediction across training fractions and trajectory-length groups.} Thin curves show individual folds and the thick curve their mean.}
\label{fig:generalization_scaling}
\end{figure}

The trajectory-length groups contain 100 short, 100 medium, and 100 long
trajectories. Prediction varies across these groups, whereas the displayed
training-fraction trend is comparatively flat. These comparisons describe
performance by trajectory structure, without separating length from domain
or label prevalence.

\subsection{Precision, Coverage, and Calibration}
\label{sec:hidden_state_generalization}
\label{app:calibration}

For READY as the positive class, precision is $TP/(TP+FP)$, recall is
$TP/(TP+FN)$, and selected coverage is $(TP+FP)/N$. If positive prevalence
is $\pi$, true-positive rate is $r$, and false-positive rate is $f$, then
\begin{equation}
 \operatorname{Precision}=\frac{\pi r}{\pi r+(1-\pi)f}.
 \label{eq:app_precision_prevalence}
\end{equation}
This follows by writing $TP=N\pi r$ and $FP=N(1-\pi)f$. In a low-prevalence
setting, false positives therefore strongly affect precision.

\begin{figure}[htbp]
\centering
\appfigure[width=.90\linewidth]{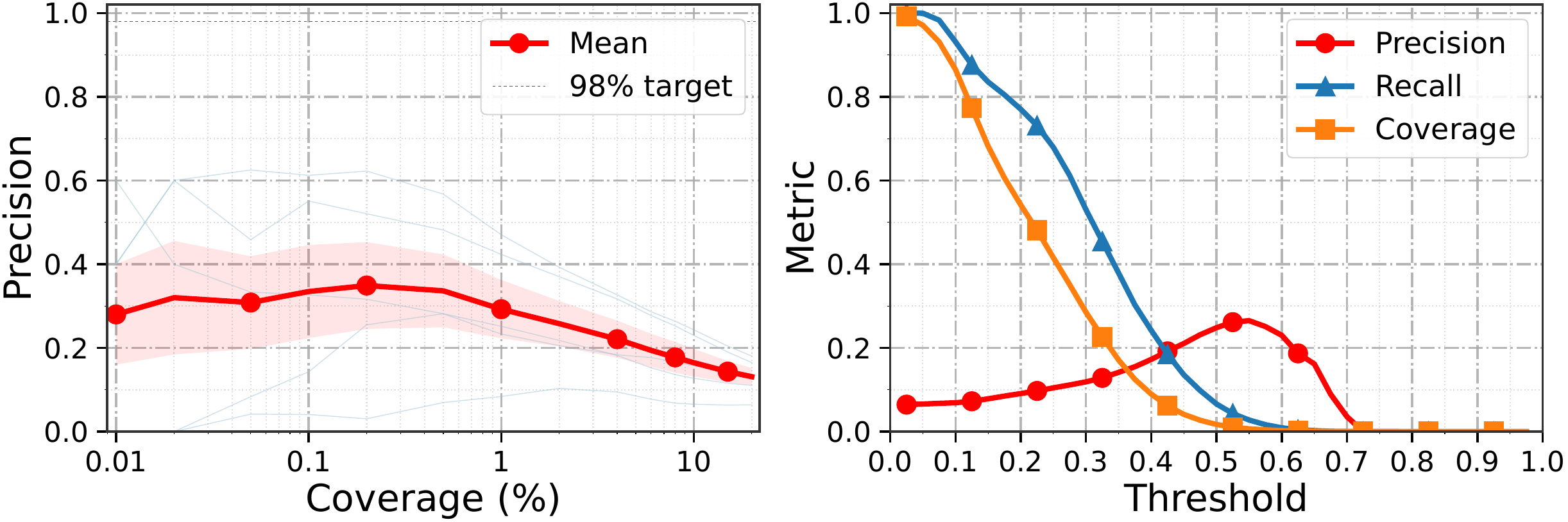}
\caption{\textbf{Held-out precision and coverage.} The 98\% line is a reference target, not an attained guarantee.}
\label{fig:safety_behavior}
\end{figure}

Raising a threshold produces nested selected sets, so coverage and FPR
cannot increase on fixed scores. Precision need not be monotone: removing
a true positive can reduce the precision of the remaining set. No
high-coverage $98\%$ precision guarantee is established by these curves.

For score bins $\mathcal B_b$, the calibration summary is
\begin{equation}
 \operatorname{ECE}=\sum_b\frac{|\mathcal B_b|}{N}
 \left|\frac1{|\mathcal B_b|}\sum_{n\in\mathcal B_b}y_n
       -\frac1{|\mathcal B_b|}\sum_{n\in\mathcal B_b}p_n\right|.
 \label{eq:app_ece}
\end{equation}
\begin{figure}[htbp]
\centering
\appfigure[width=.52\linewidth,height=.48\textheight,keepaspectratio]{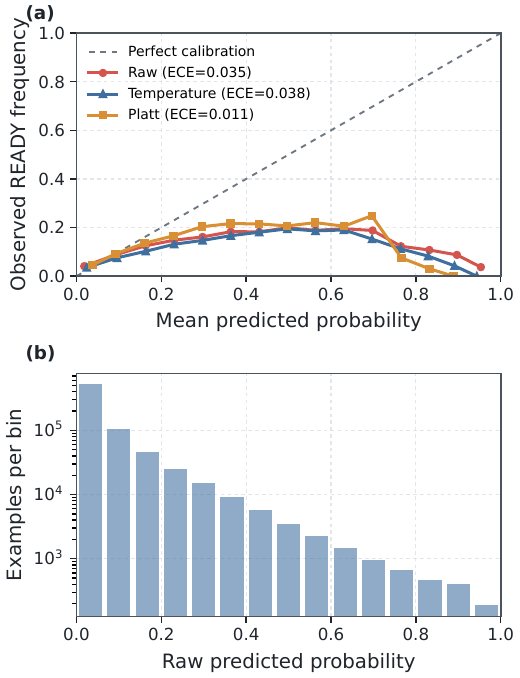}
\caption{\textbf{Calibration diagnostics.} The displayed ECE values are 0.035, 0.038, and 0.011 for raw, temperature-scaled, and Platt-calibrated scores.}
\label{fig:app_calibration}
\end{figure}

A small aggregate ECE does not imply high precision in a sparsely
populated high-score region. Calibration and selective prediction are
therefore evaluated as complementary properties.

\section{Online Compression and Analytical Properties}
\label{app:online_diagnostics}
\label{app:online_algorithm}

\subsection{Candidate Spans and Execution}

The online controller uses a frozen Qwen3.5-9B representation model with
Layer 32 last-token readouts. It concatenates the candidate-interaction
vector, the current-checkpoint vector, and the mean of the same-prefix
Top4 historical vectors, giving a 12,288-dimensional router input. The
router has one hidden layer of width 2,048 and dropout 0.3. Online
compression uses the threshold $\tau=\RouterThreshold$.

At every checkpoint, the router scores the surviving complete interactions
in the effective history. Adjacent selected interactions form maximal
spans; a KEEP interaction, protected message, or committed summary breaks
a span. A candidate $B$ is eligible when
\begin{equation}
 \min_{i\in B}p_{i,k}\ge\tau,\qquad
 |B|>\kappa,\qquad \operatorname{Tok}(B)\ge B_{\min}.
 \label{eq:app_span_gate}
\end{equation}
The compression configuration is
$(\tau,\kappa,B_{\min})=(\RouterThreshold,\SpanThreshold,\SourceThreshold)$,
as specified in Section~\ref{sec:compression_spans}. The same execution
rule is used for Full260, Eval40, and the cross-model comparisons.

Each eligible span is summarized into facts, constraints, locations,
confirmed conclusions, failure causes, and unresolved items useful to
later actions. Replacements are accepted only if the complete serialized
history remains protocol valid and becomes shorter. Later routing uses
the resulting history, with no external recovery of removed originals.

\begin{algorithm}[htbp]
\caption{One checkpoint of \method{}}
\label{alg:statecomp}
\KwRequire{Valid effective history $H_k$; frozen model $F_\phi$; trained router
$g_\theta$; threshold $\tau$; span threshold $\kappa$; source budget $B_{\min}$.}
\KwEnsure{A valid history for the next agent request.}
Preserve $H_k$ as the fallback history\;
Construct bounded inputs for surviving complete raw interactions\;
Extract features and compute $p_{i,k}$\;
\If{feature extraction or routing fails}{\Return $H_k$\;}
Form maximal eligible spans satisfying Equation~\eqref{eq:app_span_gate}\;
Initialize $\widetilde H_k\gets H_k$\;
\ForEach{eligible span $B$, in reverse history order}{
 Generate a summary $m_B$ from $B$\;
 \If{summary generation succeeds}{
  Construct a trial history by replacing $B$ in $\widetilde H_k$ with $m_B$\;
  \If{the trial is protocol valid and shorter than $\widetilde H_k$}{
   Accept the trial as the new $\widetilde H_k$ and record source identifiers\;
  }
 }
}
\If{the complete proposed history passes final protocol checks}{
 Commit $\widetilde H_k$\;
 \Return $\widetilde H_k$\;
}
\Return $H_k$\;
\end{algorithm}

\subsection{Structural Preservation and Call Bounds}

\begin{scproposition}{Valid and non-expanding updates}
\label{prop:app_atomic}
Suppose the input history is protocol valid, candidate spans contain
complete interactions, and each accepted replacement is protocol valid
and shorter under the same serialization and tokenizer. A checkpoint
returns a valid history no longer than its input. It is strictly shorter
when a nonempty replacement transaction is committed.
\end{scproposition}
\begin{scproof}
Let $H^{(0)}=H_k$ and let $H^{(j)}$ be the trial after the $j$th accepted
replacement. The acceptance rule gives
\begin{equation}
 \operatorname{Valid}(H^{(j)})=1,\qquad
 \operatorname{Tok}(H^{(j)})<\operatorname{Tok}(H^{(j-1)}).
\end{equation}
Induction proves validity and decreasing length for every accepted trial.
A rejected proposal leaves the current trial unchanged. The final
transaction either returns the last valid trial or the unchanged input.
This proves the claim. It is a property of the compression update;
subsequent interactions can increase history length again.
\end{scproof}

For a nonnegative integer $\kappa$, if a checkpoint has
$n_{\mathrm{sel}}$ selected interactions, each eligible
span consumes at least $\kappa+1$ disjoint interactions. Consequently,
\begin{equation}
 J_{\mathrm{eligible}}\le
 \left\lfloor\frac{n_{\mathrm{sel}}}{\kappa+1}\right\rfloor.
 \label{eq:sc_span_call_bound}
\end{equation}
This bounds summarization proposals under the one-call-per-span algorithm;
it does not count transport retries.

\subsection{Interaction Errors and Summary Fidelity}

\begin{scproposition}{Conditional span-error decomposition}
\label{prop:sc_span_risk}
For a selected span $B$ and available history $H$, let $E_i$ be the event
that interaction $i\in B$ is actually KEEP, and suppose
$\Pr(E_i\mid H,B)\le\epsilon_i$. Let $Q_B$ denote a harmful joint
replacement despite all selected interactions having correct READY labels.
Suppose
$\Pr(Q_B\mid H,B,\bigcap_{i\in B}E_i^c)\le\delta_B$ whenever this
conditioning event has positive probability. Then
\begin{equation}
 \Pr\!\left(\bigcup_{i\in B}E_i\;\cup\;
 \left[Q_B\cap\bigcap_{i\in B}E_i^c\right]\middle|H,B\right)
 \le \min\!\left\{1,\sum_{i\in B}\epsilon_i+\delta_B\right\}.
 \label{eq:app_span_union}
\end{equation}
\end{scproposition}
\begin{scproof}
The two bracketed failure cases are disjoint. By the union bound, the
probability of at least one incorrect READY label is at most
$\sum_i\epsilon_i$. The second case has probability
\begin{equation}
 \Pr\!\left(\bigcap_iE_i^c\middle|H,B\right)
 \Pr\!\left(Q_B\middle|H,B,\bigcap_iE_i^c\right)\le\delta_B.
\end{equation}
If the first factor is zero, this case has zero probability. Adding the
bounds and using that a probability cannot exceed one proves the result.
Independence between interaction errors is not required.
\end{scproof}

The term $\delta_B$ includes summary omissions and dependencies between
joint replacements. For example, replacing both copies of a necessary
identifier can lose information even when either copy would be replaceable
alone. The quantities in the bound are assumptions, not calibrated error
estimates obtained from aggregate precision. Minimum span size controls
execution overhead, not semantic safety by itself.

\subsection{Amortized Tokens and Prefix Caching}

\begin{scproposition}{Token break-even condition}
\label{prop:sc_amortization}
Fix an otherwise unchanged sequence of requests and outputs. Suppose a
committed replacement saves $\Delta_r$ serialized input tokens at each
subsequent request $r$, and its auxiliary summary call costs
$C_{\mathrm{sum}}$ input-plus-output tokens. Its net token saving is
\begin{equation}
 \Delta T=\sum_{r=1}^R\Delta_r-C_{\mathrm{sum}}.
 \label{eq:app_amortization}
\end{equation}
For a constant saving $\Delta>0$, the saving is positive exactly when
$R\Delta>C_{\mathrm{sum}}$.
\end{scproposition}
\begin{scproof}
Under the fixed request sequence, all unchanged input and output terms
cancel between the two accounts. The remaining input differences sum to
$\sum_r\Delta_r$, and the additional summary call contributes
$C_{\mathrm{sum}}$ only to the compressed account. Subtraction gives the
identity and the break-even condition.
\end{scproof}

Actual online runs can change actions, retries, and termination times, so
complete-task token use is measured separately. Monetary cost also depends
on cached-input, uncached-input, and output prices.

By Proposition~\ref{prop:app_prefix}, tokens before the first rewritten
position keep the same causal prefix. Tokens at and after that position
generally require recomputation. A span gate can reduce rewrite frequency,
but does not itself guarantee a larger cache-hit rate or an end-to-end
latency gain.

\section{Benchmark Protocol and Supplementary Results}
\label{app:online_results}

\subsection{Task Sets and Metrics}

WorkBuddyBench contains 80 Code, 50 Office, 60 Security, and 70 Web tasks.
Eval40 contains ten tasks from each domain. With task reward $r_t$,
\begin{equation}
\overline r=\frac1N\sum_{t=1}^Nr_t
=\sum_d\frac{N_d}{N}\overline r_d.
 \label{eq:app_reward}
\end{equation}
Tables in the main text multiply reward by 100; the detailed tables below
use the original zero-to-one scale. The overall score is the task-weighted
aggregate used in this paper. Failed task outcomes remain in fixed-set
comparisons.

For disjoint agent and summary request sets,
\begin{equation}
 T_{\mathrm{total}}=
 \sum_{r\in\mathcal R_{\mathrm{agent}}}(I_r+O_r)
 +\sum_{r\in\mathcal R_{\mathrm{summary}}}(I_r+O_r).
 \label{eq:app_token_total}
\end{equation}
Cached input and reasoning output are already included in the respective
input and output totals. Local representation computation is reported
separately. Token reductions are not assumed to equal monetary savings.

\subsection{Full260 Results}

\begin{table}[htbp]
\centering
\caption{\textbf{Full260 task rewards.} Differences use the original reward scale.}
\label{tab:app_full260_rewards}
\apptabsetup
\appfitbox{%
\begin{tabular}{l|rrrr}
\Xhline{1.1pt}
\rowcolor{TableHeader}
\textbf{Domain} & \textbf{Tasks} & \textbf{Baseline} & \textbf{\method{}} & \textbf{Difference}\\
\Xhline{0.8pt}
Code & 80 & 0.769865 & 0.771199 & +0.001334\\
\rowcolor{TableStripe}
Office & 50 & 0.818394 & 0.827696 & +0.009302\\
Security & 60 & 0.477555 & 0.473065 & -0.004490\\
\rowcolor{TableStripe}
Web & 70 & 0.721429 & 0.731429 & +0.010000\\
\hline\rowcolor{TableOurs}
Total & 260 & 0.698701 & 0.702556 & +0.003855\\
\Xhline{1.1pt}
\end{tabular}}
\end{table}

\begin{table}[htbp]
\centering
\caption{\textbf{Full260 token consumption.} Agent and summary columns include input and output tokens.}
\label{tab:full260_tokens}
\apptabsetup
\appfitbox{%
\begin{tabular}{l|rrrrr}
\Xhline{1.1pt}
\rowcolor{TableHeader}
\textbf{Domain} & \textbf{Baseline} & \textbf{Agent} & \textbf{Summary} & \textbf{Combined} & \textbf{Reduction}\\
\Xhline{0.8pt}
Code & 109,624,178 & 66,919,305 & 77,381 & 66,996,686 & 38.8851\%\\
\rowcolor{TableStripe}
Office & 59,924,344 & 26,475,291 & 43,653 & 26,518,944 & 55.7460\%\\
Security & 358,590,823 & 159,586,199 & 267,966 & 159,854,165 & 55.4216\%\\
\rowcolor{TableStripe}
Web & 170,027,887 & 79,703,907 & 165,664 & 79,869,571 & 53.0256\%\\
\hline\rowcolor{TableOurs}
Total & 698,167,232 & 332,684,702 & 554,664 & 333,239,366 & 52.2694\%\\
\Xhline{1.1pt}
\end{tabular}}
\end{table}

Combined tokens decrease from 698,167,232 to 333,239,366, corresponding
to $52.2694\%$. Mean reward changes from 0.698701 to 0.702556. Security
reward decreases slightly while the other three domain rewards increase.
These are observed task-level aggregates.

\subsection{Eval40 Controls and Summary Overhead}

The following comparisons use the same fixed 40 tasks, with ten tasks
per domain. \method{} uses the compression configuration in
Section~\ref{sec:compression_spans}. Summary tokens are reported alongside
total tokens to show the auxiliary cost of compression.

\begin{table}[htbp]
\centering
\caption{\textbf{Eval40 task rewards.} Each domain contains ten tasks; Overall covers all 40 tasks.}
\label{tab:app_eval40_domains}
\apptabsetup
\appfitbox{%
\begin{tabular}{l|rrrrr}
\Xhline{1.1pt}
\rowcolor{TableHeader}
\textbf{Method} & \textbf{Code} & \textbf{Office} & \textbf{Security} & \textbf{Web} & \textbf{Overall}\\
\Xhline{0.8pt}
Baseline & 0.7293 & 0.8164 & 0.4458 & 0.7100 & 0.6754\\
\hline\rowcolor{TableOurs}
\method{} & 0.7603 & 0.8412 & 0.5101 & 0.7500 & 0.7154\\
Sliding Window $K=5$ & 0.6681 & 0.6743 & 0.3232 & 0.5900 & 0.5639\\
\rowcolor{TableStripe}
Sliding Window $K=10$ & 0.7437 & 0.7305 & 0.4007 & 0.6100 & 0.6212\\
Sliding Window $K=20$ & 0.7931 & 0.8511 & 0.4028 & 0.7000 & 0.6867\\
\rowcolor{TableStripe}
Periodic Summary $n=3$ & 0.6544 & 0.6766 & 0.2632 & 0.8266 & 0.6052\\
Periodic Summary $n=5$ & 0.8360 & 0.7574 & 0.4003 & 0.8132 & 0.7017\\
\rowcolor{TableStripe}
PACE & 0.7038 & 0.5575 & 0.4353 & 0.6402 & 0.5842\\
\Xhline{1.1pt}
\end{tabular}}
\end{table}

\begin{table}[htbp]
\centering
\caption{\textbf{Eval40 total tokens and summary overhead.} Counts are in millions. Summary tokens are included in the total; totals follow the precision of the main-text comparison.}
\label{tab:eval40_aux}
\apptabsetup
\begin{tabular}{l|rr}
\Xhline{1.1pt}
\rowcolor{TableHeader}
\textbf{Method} & \textbf{Total tokens (M)} & \textbf{Summary tokens (M)}\\
\Xhline{0.8pt}
Baseline & 75.50 & 0\\
\hline\rowcolor{TableOurs}
\method{} & 43.76 & 0.077\\
Sliding Window $K=5$ & 118.08 & 0\\
\rowcolor{TableStripe}
Sliding Window $K=10$ & 79.39 & 0\\
Sliding Window $K=20$ & 80.82 & 0\\
\rowcolor{TableStripe}
Periodic Summary $n=3$ & 88.57 & 4.50\\
Periodic Summary $n=5$ & 94.85 & 3.01\\
\rowcolor{TableStripe}
PACE & 44.12 & 4.82\\
\Xhline{1.1pt}
\end{tabular}
\end{table}

Using the unrounded recorded token totals, the token reduction is
\begin{equation}
 100\left(1-\frac{43{,}760{,}684}{75{,}496{,}039}\right)\approx42.04\%.
 \label{eq:sc_eval40_reduction}
\end{equation}
The compression run uses approximately 0.077M summary tokens, compared
with 3.01M for periodic summarization at $n=5$. The full control table
shows both the performance losses of short windows and the overhead of
repeated summarization.

\paragraph{Additional published baselines.}
\label{app:baseline_scope}
The main comparison also includes published context-management methods.
LLMLingua-2, SelfCompact, ACON-Core, Self-GC, LRE, CoMem, SAM,
SWE-Pruner, Sculptor, and ACM are evaluated on the same fixed Eval40 tasks.
Their token totals use the same accounting rule as the other Eval40 methods.
ACM uses a released policy, rather than the same API-backed acting model.

\subsection{Cross-Model Results}

All acting models are evaluated on the same fixed Eval40 tasks with the
compression rule in Section~\ref{sec:compression_spans}. Each domain
contains ten tasks, so overall reward is the equally weighted mean of
the four domain rewards. Summary usage describes the auxiliary cost of
compression.

\begin{table}[htbp]
\centering
\caption{\textbf{Cross-model Eval40 rewards.} Rewards use the zero-to-one scale. Each domain contains ten tasks.}
\label{tab:cross_model_domains}
\apptabsetup
\appfitbox{%
\begin{tabular}{ll|ccccc}
\Xhline{1.1pt}
\rowcolor{TableHeader}
\textbf{Model} & \textbf{Method} & \textbf{Code} & \textbf{Office} & \textbf{Security} & \textbf{Web} & \textbf{Overall}\\
\Xhline{0.8pt}
Hunyuan & Base & 0.5530 & 0.7924 & 0.6024 & 0.4600 & 0.6019\\
\hline\rowcolor{TableOurs}
Hunyuan & \method{} & 0.6431 & 0.8220 & 0.6014 & 0.4900 & 0.6391\\
Qwen3.8-Flash & Base & 0.7525 & 0.8342 & 0.4528 & 0.8600 & 0.7249\\
\hline\rowcolor{TableOurs}
Qwen3.8-Flash & \method{} & 0.9346 & 0.8871 & 0.4120 & 0.8800 & 0.7784\\
GLM-5.3-Flash & Base & 0.7949 & 0.6613 & 0.3467 & 0.6651 & 0.6170\\
\hline\rowcolor{TableOurs}
GLM-5.3-Flash & \method{} & 0.6615 & 0.6990 & 0.3864 & 0.8500 & 0.6492\\
MiMo-V2.5 & Base & 0.5376 & 0.7705 & 0.2204 & 0.5500 & 0.5196\\
\hline\rowcolor{TableOurs}
MiMo-V2.5 & \method{} & 0.5765 & 0.7164 & 0.2277 & 0.5800 & 0.5252\\
\Xhline{1.1pt}
\end{tabular}}
\end{table}

\begin{table}[htbp]
\centering
\caption{\textbf{Cross-model total tokens and summary overhead.} Counts are in millions. Summary tokens are included in the total.}
\label{tab:app_cross_tokens}
\apptabsetup
\begin{tabular}{ll|rr}
\Xhline{1.1pt}
\rowcolor{TableHeader}
\textbf{Model} & \textbf{Method} & \textbf{Total tokens (M)} & \textbf{Summary tokens (M)}\\
\Xhline{0.8pt}
Hunyuan & Base & 41.00 & 0\\
\hline\rowcolor{TableOurs}
Hunyuan & \method{} & 12.83 & 0.130\\
Qwen3.8-Flash & Base & 49.75 & 0\\
\hline\rowcolor{TableOurs}
Qwen3.8-Flash & \method{} & 38.23 & 0.236\\
GLM-5.3-Flash & Base & 47.58 & 0\\
\hline\rowcolor{TableOurs}
GLM-5.3-Flash & \method{} & 38.49 & 0.101\\
MiMo-V2.5 & Base & 53.55 & 0\\
\hline\rowcolor{TableOurs}
MiMo-V2.5 & \method{} & 18.91 & 0.028\\
\Xhline{1.1pt}
\end{tabular}
\end{table}

All four acting models use fewer total tokens with compression. Reward
changes and their distribution across domains remain model dependent.
The totals characterize complete-task usage; summary tokens are an
included component, not an additional charge to the reported totals.

\section{Parameter Analysis and Evaluation Tasks}
\label{app:parameter_studies}

\subsection{Operating Settings}
\label{app:config_scope}

The probability threshold $\tau$, representation window $K_{\mathrm{ctx}}$,
sliding-window baseline $K_{\mathrm{sw}}$, source-token threshold
$B_{\min}$, and span-length threshold $\kappa$ control distinct operations.
Full260, Eval40, and the cross-model comparisons use $\tau=\RouterThreshold$,
$\kappa=\SpanThreshold$, and $B_{\min}=\SourceThreshold$ tokens. Thus, a
span must contain at least four complete interactions and satisfy the
source-token threshold before summarization. The representation budget
remains at most $\StateBudget$ tokens per target view.

\subsection{Fixed Eval40 Task Identifiers}
\label{app:eval40_manifest}

The fixed task list contains ten tasks from each domain. It specifies the
benchmark subset used for the Eval40 comparisons.

\begin{table}[htbp]
\centering
\caption{\textbf{Fixed Eval40 task identifiers.}}
\label{tab:app_eval40_ids}
\apptabsetup
\appfitbox{%
\begin{tabular}{r|l|l}
\Xhline{1.1pt}
\rowcolor{TableHeader}
\textbf{No.} & \textbf{Domain} & \textbf{Task identifier}\\
\Xhline{0.8pt}
1 & Code & \texttt{\detokenize{api_contract-hard-markup_errors}}\\
\rowcolor{TableStripe}
2 & Code & \texttt{\detokenize{api_contract-hard-openapi_params}}\\
3 & Code & \texttt{\detokenize{api_contract-hard-token_errors}}\\
\rowcolor{TableStripe}
4 & Code & \texttt{\detokenize{api_contract-hard-validation_errors}}\\
5 & Code & \texttt{\detokenize{bug_fix-easy-a_crash_in_local}}\\
\rowcolor{TableStripe}
6 & Code & \texttt{\detokenize{bug_fix-easy-filtered_relation_queryset_arg}}\\
7 & Code & \texttt{\detokenize{bug_fix-easy-invalid_filterwarnings_regex_error}}\\
\rowcolor{TableStripe}
8 & Code & \texttt{\detokenize{bug_fix-medium-error_key_uses_data_key}}\\
9 & Code & \texttt{\detokenize{bug_fix-medium-errors_from_earlier_indices}}\\
\rowcolor{TableStripe}
10 & Code & \texttt{\detokenize{bug_fix-medium-incorrect_linenos_on_fstring}}\\
11 & Office & \texttt{\detokenize{analyst-forecast-extract-L3-018}}\\
\rowcolor{TableStripe}
12 & Office & \texttt{\detokenize{api-usage-explain-cli-l3-001}}\\
13 & Office & \texttt{\detokenize{board-material-update-timeline-excel}}\\
\rowcolor{TableStripe}
14 & Office & \texttt{\detokenize{calendar-dida-sync-state}}\\
15 & Office & \texttt{\detokenize{channel-period-compare-L4-017}}\\
\rowcolor{TableStripe}
16 & Office & \texttt{\detokenize{cloudagent-sdk-doc-validation-report}}\\
17 & Office & \texttt{\detokenize{contract-extract-L3-014}}\\
\rowcolor{TableStripe}
18 & Office & \texttt{\detokenize{cross-week-dashboard-migration}}\\
19 & Office & \texttt{\detokenize{crypto-backtest-chain-L4-002}}\\
\rowcolor{TableStripe}
20 & Office & \texttt{\detokenize{daily-creation-checkpoint-recovery}}\\
21 & Security & \texttt{\detokenize{agent-to-agent-injection-hard-multistep}}\\
\rowcolor{TableStripe}
22 & Security & \texttt{\detokenize{apt-multi-source-correlation-hard-multistep}}\\
23 & Security & \texttt{\detokenize{bb-bin-dns-parse-010}}\\
\rowcolor{TableStripe}
24 & Security & \texttt{\detokenize{bb-bin-firmware-audit-007}}\\
25 & Security & \texttt{\detokenize{bb-bin-format-log-004}}\\
\rowcolor{TableStripe}
26 & Security & \texttt{\detokenize{bb-bin-int-length-005}}\\
27 & Security & \texttt{\detokenize{bb-bin-ipc-cache-001}}\\
\rowcolor{TableStripe}
28 & Security & \texttt{\detokenize{bb-bin-media-parse-008}}\\
29 & Security & \texttt{\detokenize{bb-bin-oob-read-003}}\\
\rowcolor{TableStripe}
30 & Security & \texttt{\detokenize{bb-bin-parse-crash-006}}\\
31 & Web & \texttt{\detokenize{animated-explainer-L3-028}}\\
\rowcolor{TableStripe}
32 & Web & \texttt{\detokenize{atmosphere-game-L4-035}}\\
33 & Web & \texttt{\detokenize{blog-editor-draft-recovery-L4-059}}\\
\rowcolor{TableStripe}
34 & Web & \texttt{\detokenize{browser-clipper-extension-L4-005}}\\
35 & Web & \texttt{\detokenize{canvas-webgl-scene-L4-026}}\\
\rowcolor{TableStripe}
36 & Web & \texttt{\detokenize{chart-generation-L2-025}}\\
37 & Web & \texttt{\detokenize{checkout-incident-analysis-L4-049}}\\
\rowcolor{TableStripe}
38 & Web & \texttt{\detokenize{city-article-theme-variants-L4-066}}\\
39 & Web & \texttt{\detokenize{claims-drawer-state-review-report-L3-071}}\\
\rowcolor{TableStripe}
40 & Web & \texttt{\detokenize{cohort-retention-dashboard-L4-054}}\\
\Xhline{1.1pt}
\end{tabular}}
\end{table}

\clearpage
% End of replacement appendix.
\end{document}